\documentclass[pdflatex,sn-mathphys-num]{sn-jnl}

\makeatletter
\renewcommand\paragraph[1]{%
  \par\medskip\noindent
  {\normalfont\normalsize\bfseries #1}\quad\ignorespaces}
\makeatother

\usepackage{graphicx}
\usepackage{multirow}
\usepackage{amsmath,amssymb,amsfonts}
\usepackage{amsthm}
\usepackage{mathrsfs}
\usepackage[title]{appendix}
\usepackage{xcolor}
\usepackage{textcomp}
\usepackage{manyfoot}
\usepackage{booktabs}
\usepackage{algorithm}
\usepackage{algorithmicx}
\usepackage{algpseudocode}
\usepackage{listings}

\usepackage{enumitem}
\setlist{leftmargin=2.5em}

\usepackage[autostyle]{csquotes}
\MakeOuterQuote{"}

\usepackage{cleveref}
\crefname{figure}{Figure}{Figures}
\crefname{table}{Table}{Tables}
\crefname{section}{Section}{Sections}
\crefname{appendix}{Appendix}{Appendices}

\title[Article Title]{OOD Detection for EEG-based Machine Learning in High-Risk Environments}

\author*[1]{\fnm{Philipp} \sur{Bomatter}}\email{philipp.bomatter@ed.ac.uk}
\author[1]{\fnm{Henry} \sur{Gouk}}\email{henry.gouk@ed.ac.uk}
\affil[1]{
    \orgdiv{School of Informatics},
    \orgname{University of Edinburgh},
    \orgaddress{
        \country{United Kingdom}%
        }%
}

\abstract{
    Machine learning models for electroencephalography (EEG) analysis show great promise across a wide range of applications, but their deployment in high-risk domains is hindered by their vulnerability to distribution shifts. Encountering out-of-distribution (OOD) data can lead to catastrophic, overconfident predictive failures. While OOD detection methods can mitigate these risks, they remain heavily under-explored for EEG. Moreover, evaluations in the broader literature typically evaluate OOD detection performance in isolation, ignoring their practical impact on downstream applications. To bridge this gap, we introduce a benchmark for EEG OOD detection, evaluate a broad range of methods, and furthermore evaluate their value in two clinical downstream prediction task. Our results disentangle OOD detection and model uncertainty estimation capabilities, which are frequently conflated in the literature, provide actionable insights about the current state of the art for EEG OOD detection and model uncertainty estimation, and demonstrate how complementary methods for both aspects can be combined to form a robust safety net for the deployment of EEG-based machine learning models in real-world applications.
}


\begin{document}

\maketitle

\section{Introduction}
\label{sec:introduction}

    Machine learning (ML) provides a promising avenue for supporting and enhancing the analysis of electroencephalography (EEG). However, especially in high-risk contexts like clinical applications, the use of machine learning-based EEG biomarkers is hindered by their vulnerability to distribution shifts. Machine learning models can fail catastrophically (and with high confidence) when they are presented with data that differs substantially from the data they were trained on, i.e. out-of-distribution (OOD) data \cite{nguyen_deep_2014}. Particularly for EEG, there are ample ways in which such shifts can be introduced, including different recording hardware, (accidental) mismatches in the preprocessing (e.g. filter settings or re-referencing), or differences in the patient population. While specific quality controls for some of these can be designed, part of the problem is that there can be unknown shifts that one does not expect.
    Furthermore, while for example individual differences between patients can be the cause of distribution shifts \cite{bomatter_is_2025}, only deploying a model for patients that the model has seen during training is often not an option. In such cases, it may not be clear a priori whether the EEG data from a new patient is similar enough to the data seen during training. 
    
    OOD detection can mitigate these risks by enabling a system to abstain from making a prediction or defer to a human expert when a case is likely out of distribution.
    Current OOD detection methods can broadly be categorized into two paradigms: discriminative and generative. Approaches such as Maximum Softmax Probability (MSP) \cite{hendrycks_baseline_2018}, Energy scores \cite{liu_energy-based_2020}, Out-of-Distribution Detector for Neural Networks (ODIN) \cite{liang_enhancing_2020}, and Activation Shaping (ASH) \cite{djurisic_extremely_2023} directly operate on the activations or outputs of a (discriminative) classification model. Conversely, methods based on a generative model---including likelihood estimates, Typicality \cite{nalisnick_detecting_2019}, Density of State Estimation (DoSE) \cite{morningstar_density_2021}, and Signal in the Noise (SITN) \cite{bomatter_signal_2026}---attempt to explicitly model the underlying distribution of the training data. While these methods have seen extensive development and benchmarking in domains like computer vision, their application to EEG data remains heavily under-explored.
    Furthermore, the standard evaluation paradigm in the broader OOD detection literature suffers from a significant disconnect from clinical utility. The vast majority of existing work evaluates OOD detection methods strictly in isolation, assessing only how well a method can separate an in-distribution dataset from a distinctly different out-of-distribution dataset (e.g. separating CIFAR-10 images from SVHN images). Crucially, such evaluations fail to reflect the downstream impact of OOD detection: when deploying a system for clinical diagnosis, the ultimate goal is not to merely detect OOD samples, but to understand how flagging such samples improves the reliability of the diagnosis.

    To bridge this gap, this work presents a systematic evaluation of OOD detection for EEG, explicitly linking detection capabilities to downstream performance in clinical prediction tasks. We design a suite of practically relevant perturbations with different levels of severity and use them to benchmark a broad set of OOD detection methods spanning the aforementioned approaches based on both discriminative and generative models.
    In line with a recent theoretical critique and position paper \cite{li_position_2025, li_out--distribution_2025}, our results disentangle two fundamentally distinct concepts that are frequently conflated in the literature: true OOD detection vs model uncertainty. Across perturbations, methods based on discriminative models were largely incapable of OOD detection, while methods based on generative models achieved considerably better results, with increasing performance for higher perturbation severities. However, in downstream clinical evaluations---where model errors stem from both OOD data and difficult in-distribution (ID) cases---we find that methods from both paradigms effectively predict model performance. While generative models recognise unfamiliar data, discriminative methods can serve as a valuable proxy for model uncertainty. Ultimately, we show that these complementary signals can be combined to form a more robust safety net for the deployment of EEG-based machine learning models in clinical and other high-risk applications.

    \vspace{0.25cm} \noindent
    In summary, our main contributions are as follows:
    
    \begin{itemize}
        \item We present an OOD detection evaluation framework for EEG that closes several gaps in the field. The framework comprises a suite of perturbations for the controlled creation of OOD EEG data, as well as investigations of downstream impacts in two clinically relevant prediction tasks.
        
        \item We benchmark a broad range of OOD detection methods, including both discriminative and generative approaches, the latter of which have not previously been probed in an EEG context, but achieved substantially better performance.
        
        \item We present strong empirical evidence that clearly disentangles the often conflated concepts of OOD detection and model uncertainty, a finding and experimental setup that is of broader interest to the OOD detection community beyond EEG.
        
        \item Our results provide actionable guidance on the current state of the art for EEG OOD detection and show how it can be combined with uncertainty estimation for the safer deployment of machine learning models in high-risk EEG applications.
    \end{itemize}
    \vspace{0.2cm}

\section{Related Work}
    \paragraph{OOD Detection for EEG-based Machine Learning}
        The EEG-specific literature on OOD detection is limited.
        In a recent preprint, \citeauthor{mulder_challenge_2026} benchmarked several discriminative OOD detection methods in a motor imagery setting and found them to be largely ineffective \cite{mulder_challenge_2026}.
        \citeauthor{liu_temporal_2026} proposed a temporal OOD detection method for EEG-based brain-computer interfaces (BCIs) and compared it against a range of standard discriminative OOD detection methods \cite{liu_temporal_2026}. Our study investigates a more general setting than the latter and is focused on a clinical context rather than BCIs. Furthermore, while our results largely concur with their findings on discriminative methods, our experiments also include methods based on generative models, an important addition given that they displayed much better OOD detection capabilities.
        Concurrent work by \citeauthor{tveter_uncertainty_2026} also introduced a range of perturbations to create EEG data with controlled distribution shifts \cite{tveter_uncertainty_2026}. However, their work is focused on the comparison of different ensemble learning methods and their robustness to distribution shifts rather than OOD detection.

    \paragraph{OOD Detection Benchmarks in Other Domains}
        Outside of EEG, the OOD detection literature is particularly advanced in computer vision. Notable works that have focused on the evaluation of different OOD detection methods are the OpenOOD benchmarks \cite{yang_openood_2022, zhang_openood_2024}, although many methods papers also include comprehensive comparisons against other methods \cite{hendrycks_baseline_2018, liu_energy-based_2020, liang_enhancing_2020, djurisic_extremely_2023, morningstar_density_2021, bomatter_signal_2026}.
        Less mature, but more closely related to EEG, is the field of OOD detection for (more general) time series data. Within this field, the OOD detection benchmark for time series data by \citeauthor{gungor_ts-ood_2025} is most relevant to our work \cite{gungor_ts-ood_2025}. However, like EEG-specific prior work, their benchmarks are limited to discriminative approaches, for which they reported relatively poor performance overall.
        Furthermore, a common thread across all these works is their primary reliance on semantic shifts to define OOD data, such as reserving a subset of classes as OOD or evaluating models across disjoint datasets.
        \citeauthor{guille-escuret_expecting_2024} recently criticised these approaches, arguing that an exclusive focus on semantic shifts neglects relevant (covariate) distribution shifts encountered by real-world systems \cite{guille-escuret_expecting_2024}.
        Motivated by a practical perspective, we see the value of OOD detection in protecting against model failure, irrespective of the nature of the shift. Our work is therefore closely aligned with this view.
        Our work is also different in that we go beyond standard OOD detection evaluations to study impacts on downstream clinical applications and disentangle the concepts of true OOD detection and model uncertainty.

\section{Methods}
\label{sec:methods}

    \begin{figure}[t]
        \centering
        \includegraphics[width=\textwidth]{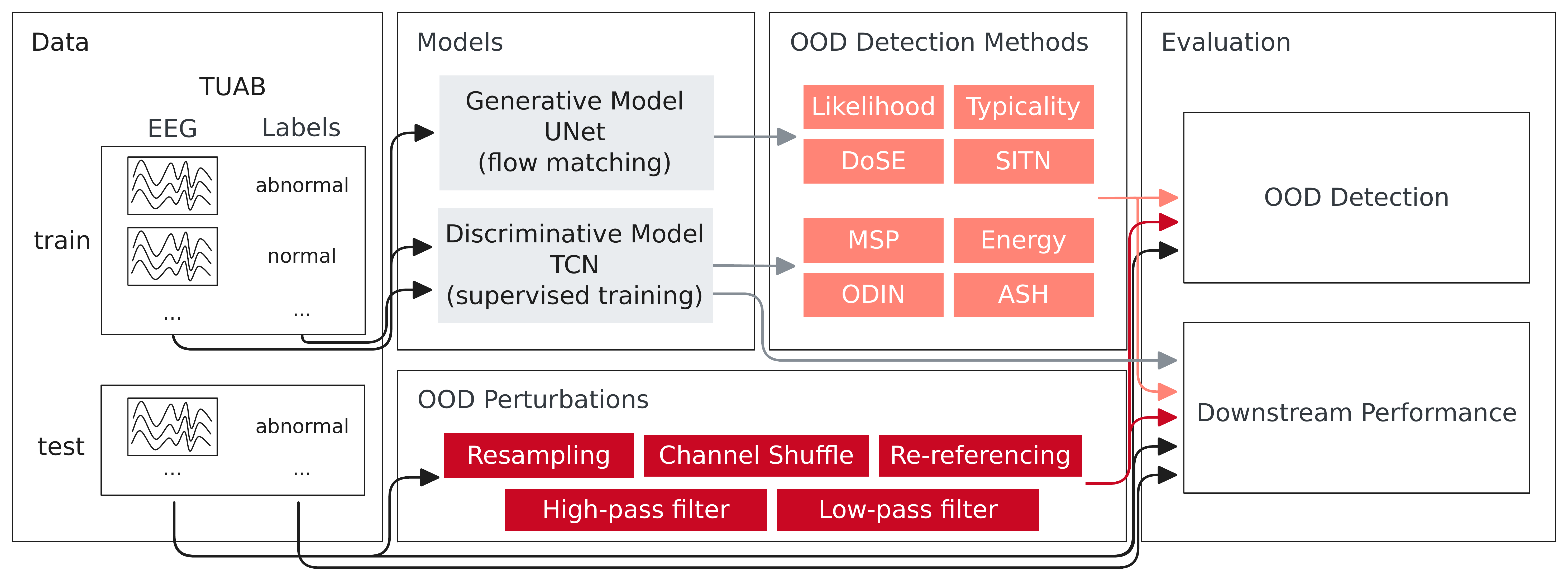}
        \caption{
            \textbf{OOD Detection Evaluation Concept.} Starting from an EEG dataset with annotations for a prediction task (here TUAB with EEG normality labels), the data is first partitioned into training and test splits. The training data is then used to train two models: a generative model trained using only the EEG without labels (here a UNet trained via flow-matching), and a discriminative model trained through supervised learning (here a TCN model). The test data then serves as in-distribution test data, for which corresponding out-of-distribution data is generated through controlled perturbations. Finally, we evaluate a range of OOD detection methods based on either the generative or discriminative model. In addition to OOD detection performance, we also evaluate its impact on downstream performance, quantifying how the prediction accuracy of the discriminative model varies across samples with different OOD scores of the different methods.
        }
        \label{fig:concept}
    \end{figure}

    \cref{fig:concept} provides an overview of the components that enable our EEG OOD detection evaluations. Starting from an EEG dataset with corresponding annotations for a (clinical) prediction task, samples are first partitioned into a training and test split. The training data is used to train two models: a generative model trained using only the EEG without labels (i.e. unsupervised), and a discriminative model trained through supervised learning. The test set then serves as in-distribution test data, for which we generate corresponding out-of-distribution data through controlled perturbations. Finally, we evaluate a range of OOD detection methods based on either the generative or discriminative model in two ways. First, we assess their OOD detection performance, i.e. how well their assigned OOD scores discriminate between the unperturbed ID test samples and their corresponding perturbed OOD versions. Second, we also evaluate the impact of OOD detection on downstream performance in the (clinical) prediction task, quantifying how the prediction accuracy of the discriminative model varies across samples with different OOD scores of the different methods.
    
    Detailed descriptions for the different components are given in the following sections. We start with the EEG perturbation suite that forms the backbone of our benchmarks in \cref{sec:perturbation_suite}. \cref{sec:ood_methods} then outlines the different OOD detection methods included in our benchmarks along with 
    architecture and training details for the underlying models. Information about the datasets, prediction tasks, data splitting, and preprocessing is provided in \cref{sec:data}.

    \subsection{OOD Perturbation Suite for EEG}
    \label{sec:perturbation_suite}

\begin{figure}[t]
            \centering
            \includegraphics[width=\textwidth]{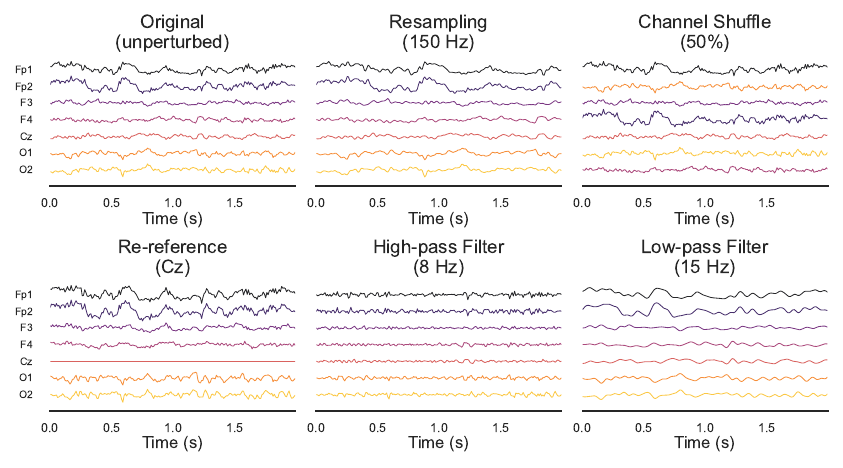}
            \caption{\textbf{Illustration of EEG OOD Perturbations.} An unperturbed (ID) EEG segment is shown (top left panel) along with corresponding OOD samples obtained through the different perturbations. For visualisation purposes, only a subset of channels are shown.}
            \label{fig:perturbations}
        \end{figure}

        We introduce a suite of controlled perturbations to create OOD samples from in-distribution test data. These transformations were designed to simulate plausible distribution shifts that may be encountered in real-world EEG data, though they are fundamentally intended to serve as a structured, reproducible, and well-controlled testbed rather than an exhaustive modelling of reality. Furthermore, the perturbations were designed to be parametrisable, which (with the exception of the re-referencing perturbation) enables fine-grained control of the severity of the distribution shift. Finally, by applying these transformations directly to the ID test set, the resulting OOD data is perfectly paired with the unperturbed baseline, providing the methodological capacity to isolate the exact effect of a shift on model performance. The perturbations are illustrated in \cref{fig:perturbations} and described in the following paragraphs.

        \paragraph{Sampling Rate}
            The signal is resampled to a higher sampling rate (using bandlimited sinc interpolation via the torchaudio library) and then cropped to maintain the same number of samples. From the perspective of a model that expects data to be at a lower sampling rate, perturbed samples will appear smoothed. Sampling rate mismatches are surprisingly common in practice; for instance, a common issue in the evaluation of EEG foundation models is that they are applied to data with a different sampling rate than the model was pre-trained on. We control the severity of this perturbation by varying the amount by which the sampling rate is increased compared to the 100 Hz that the model was trained on, ranging from 125 Hz over 150 Hz to 200 Hz.

        \paragraph{Channel Order}
            The ordering of a subset of EEG channels is permuted. From the perspective of a model that was trained with a fixed given channel order, the perturbation will lead to strange topographies with, for example, eye blinks appearing in presumed occipital channels. In practice, perturbations like this could be introduced by accident with some recording devices where electrodes are plugged into the amplifier individually. We control the severity of this perturbation by varying the size of the subset of randomly selected channels which are permuted. We increase the subset from 10\%, i.e. swapping just two channels in the 19-channel layout, to 50\% and then 100\%.
            
        \paragraph{Reference Scheme}
            The EEG channels are re-referenced to a different scheme. Starting from average-referenced data that is used to train the model, we re-reference to three different schemes: a monopolar vertex reference with Cz as the reference channel, a linked-temporal reference with the average of T7 and T8 as the reference, and a longitudinal double-banana bipolar montage with an additional transverse channel (T7-T8) to maintain the same channel count. Unlike the different levels for other perturbations, these three options do not have a clear ordering in terms of perturbation severity.

        \paragraph{High-pass Filter}
            The signal is filtered with an additional zero-phase Butterworth high-pass filter, attenuating low-frequency content. Starting from the unperturbed data (high-pass filtered at 1 Hz), we vary the severity of this perturbation by increasing the cutoff frequency starting from 2 Hz to 4 Hz and 8 Hz, progressively removing more low-frequency activity.

        \paragraph{Low-pass Filter}
            Analogous to the high-pass filter perturbation described above, the signal is filtered with a low-pass filter. Starting from the unperturbed data (low-pass filtered at 50 Hz), we vary the severity of this perturbation by lowering the cutoff frequency starting from 45 Hz to 30 Hz and 15 Hz, progressively removing more high-frequency activity.

    \subsection{OOD Detection Methods}
    \label{sec:ood_methods}

        Our benchmarks include popular OOD detection methods grouped into two paradigms: methods based directly on discriminative classification models and methods based on generative models.

        \subsubsection{Discriminative-model-based Methods}
            In the following, we outline the OOD detection methods included in our benchmarks that are directly based on the discriminative model used for the given classification task. As the underlying discriminative model for all methods, we use the TCN architecture originally proposed by \citeauthor{bai_empirical_2018} \cite{bai_empirical_2018}, which has since been adapted and widely used for EEG data \cite{gemein_machine-learning-based_2020} (see \cref{appendix:training_details} for hyperparameters and training details).
        
            \paragraph{Maximum Softmax Probability (MSP)}
                MSP uses the maximum probability from the network's final softmax layer as a confidence score \cite{hendrycks_baseline_2018}. This is based on the assumption that the model will output lower maximum probabilities (reflecting higher uncertainty) when presented with OOD data compared to ID data.
            
            \paragraph{Energy Score}
                Instead of operating on the softmax outputs, which are normalised to sum to one, \citeauthor{liu_energy-based_2020} proposed computing an energy score directly from the model's unnormalised logits \cite{liu_energy-based_2020}. OOD samples are expected to produce higher energy scores.
            
            \paragraph{Out-of-Distribution Detector for Neural Networks (ODIN)}
                ODIN builds upon the standard MSP approach through two key modifications: applying temperature scaling to the model's logits prior to the softmax computation, and adding gradient-based perturbations to the input data that are expected to have a stronger impact on ID than OOD data \cite{liang_enhancing_2020}. Together, these modifications aim to improve the separability of ID and OOD samples with the resulting score. We adopt the hyperparamters reported in the original publication for the main results (T=1000, $\epsilon$=0.0014). We did not perform hyperparameter tuning because we do not assume access to OOD data in our experiments. Furthermore, the original publication suggests relatively good hyperparameter transferability. 
            
            \paragraph{Activation Shaping (ASH)}
                \citeauthor{djurisic_extremely_2023} introduced an activation shaping (ASH) method that intervenes in the forward pass of a model by pruning and transforming activations before they reach the final classification head, after which the energy score is computed on the resulting logits \cite{djurisic_extremely_2023}. We specifically use the ASH-S version with a pruning percentage of 65\%. While tuning of this hyperparameter would require access to OOD data, which we do not assume in our benchmarks, the authors report relatively stable performance across different parameter values in the original publication.

        \subsubsection{Generative-model-based Methods}
            In the following, we outline the OOD detection methods included in our benchmarks that are based on a generative model.
            As the underlying generative model for all methods, we use the UNet architecture following \citeauthor*{ho_denoising_2020} \cite{ho_denoising_2020}, as implemented by \citeauthor*{dhariwal_diffusion_2021} \cite{dhariwal_diffusion_2021}, and train it using the flow matching objective by \citeauthor*{lipman_flow_2023} \cite{lipman_flow_2023} (see \cref{appendix:training_details} for details). We use Euler's method (with 100 timesteps), implemented in the library by \citeauthor*{lipman_flow_2024} \cite{lipman_flow_2024}, as ODE solver to obtain the likelihoods and noise samples required by the different methods.
        
            \paragraph{Log-likelihood}
                The log-likelihood is evaluated via the instantaneous change of variables formula, calculated as the sum of the log-probability of the latent representation and the negative integral of the vector field's divergence over the flow trajectory. Samples with low likelihood under the generative model are flagged as OOD, reflecting the assumption that the model assigns higher probability densities to ID data.
            
            \paragraph{Typicality}
                Typicality was introduced by \citeauthor{nalisnick_detecting_2019} to correct for a failure mode of likelihoods in high dimensions where regions of high probability \emph{density} can lie outside of the so-called typical set that contains almost all probability \emph{mass} \cite{nalisnick_detecting_2019}. It is defined as the absolute difference between the negative log-likelihood and an estimate of the entropy. We estimate the entropy as the average negative log-likelihood over the (in-distribution) validation data, following experiments in \cite{bomatter_signal_2026} that showed negligible differences to estimation on the training data as in the original publication. Furthermore, we use the single-sample version as in \cite{morningstar_density_2021}.
            
            \paragraph{Density of State Estimation (DoSE)}
                DoSE measures the empirical density of multiple statistics computed on in-distribution data (as for Typicality, we use (in-distribution) validation data following experiments in \cite{bomatter_signal_2026}) \cite{morningstar_density_2021}.
                We specifically use the $\text{DoSE}_{\text{KDE}}$ version for flow-based models, which is based on three statistics: the log-likelihood, the log-probability of the latent representation, and the log-determinant of the Jacobian (corresponding to the integral of the vector field's divergence for the continuous-time flow matching model used in our experiments).
                A 1D Kernel Density Estimator (KDE) is fit to each statistic and the final OOD score is computed by summing the log-probabilities from the individual KDEs. As in the original publication, the default SciPy KDE implementation with a Gaussian kernel and automatic determination of the bandwidth parameter through Scott's rule is used, and following \cite{bomatter_signal_2026}, the KDEs are fit on a random subset of 10,000 samples for computational efficiency.
            
            \paragraph{Signal in the Noise (SITN)}
                SITN uses the diffeomorphic properties of continuous normalising flows to detect OOD samples by checking if their corresponding noise samples (obtained through backwards integration along the probability flow ODE) are consistent with the Gaussian prior used during training \cite{bomatter_signal_2026}.

    \subsection{Data}
    \label{sec:data}

        Experiments were conducted on two large datasets (TUAB, CAUEEG) with annotations for clinical prediction tasks. With more than 1{,}000 participants each, these datasets are among the largest EEG datasets available for research purposes, providing us with the necessary sample size for robust training and benchmarking of machine learning methods.

        \paragraph{TUAB}
            The Temple University Hospital Abnormal (TUAB) EEG Corpus \cite{lopez_automated_2015, diego_automated_2017} comprises a demographically balanced and curated subset of more than 2{,}300 participants from the TUH EEG Corpus \cite{obeid_temple_2016}. The data was collected in a hospital environment and recordings were annotated by board certified neurologists for the normality of the EEG according to standardised criteria \cite{rubin_clinical_2021}. We specifically used version 3.0.1 of the dataset and followed the data curation and splitting in \cite{bomatter_is_2025}, which respects the official train-test split and additionally defines a validation split with 10\% of the training data. All splitting is performed on the participant level, i.e. all samples of a given participant are assigned to the same split. Data access requests should be directed to the dataset authors.

        \paragraph{CAUEEG}
            The Chung-Ang University Hospital EEG (CAUEEG) dataset \cite{kim_deep_2023} includes a subset of 1{,}122 participants to which diagnostic labels for the categories normal, mild cognitive impairment (MCI), or dementia were assigned by neurologists based on a neuropsychological examination.
            We used the official train-validation-test split without overlapping participants. Data access for academic and research purposes can be requested from the dataset authors.

        \paragraph{Preprocessing}
            The data preprocessing comprised the following steps: selection of a fixed channel subset (the 19 channels in the 10-20 system), band-pass filtering to 1-50 Hz, resampling to 100 Hz, and re-referencing to average reference. Finally, non-overlapping epochs with 2s duration were used.

\section{Results}
\label{sec:results}

    \subsection{OOD Detection Performance}
    \label{sec:results_ood_detection}

        \cref{tab:ood_auroc_tuab} shows the Area Under the Receiver Operating Characteristic (AUROC) curve for the OOD detection performance across the different perturbations and severity levels for all methods on the TUAB dataset. Models for the discriminative and generative methods were trained on the train split of the (unperturbed) dataset and OOD detection methods were then evaluated by using the unperturbed test split of the dataset as ID samples and the perturbed test data as OOD samples.

        \begin{table*}[h]
            \centering
            \small
            \setlength{\tabcolsep}{3pt}
            \caption{
                \textbf{OOD detection performance on TUAB.} Performance is reported in terms of AUROC and bold values highlight the best result for each perturbation and severity level. Discriminative-model-based methods (MSP, Energy, ODIN, ASH) perform largely at chance level, whereas generative-model-based methods achieve considerably better OOD detection with increasing performance for higher severity levels. SITN consistently achieves the best performance.
            }
            \label{tab:ood_auroc_tuab}
            \begin{tabular}{ll cccccccc}
                \toprule
                \textbf{Perturbation} & \textbf{Severity} & \textbf{MSP} & \textbf{Energy} & \textbf{ODIN} & \textbf{ASH} & \textbf{LL} & \textbf{Typ.} & \textbf{DoSE} & \textbf{SITN} \\
                \midrule
                \multirow{3}{*}{Resampling}
                & 125 Hz & 0.517 & 0.500 & 0.514 & 0.533 & 0.369 & 0.559 & 0.556 & \textbf{0.947} \\
                & 150 Hz & 0.457 & 0.479 & 0.453 & 0.534 & 0.275 & 0.646 & 0.647 & \textbf{0.981} \\
                & 200 Hz & 0.383 & 0.451 & 0.377 & 0.522 & 0.143 & 0.816 & 0.826 & \textbf{0.989} \\
                \midrule
                \multirow{3}{*}{Channel Shuffle}
                & 10\%  & 0.549 & 0.494 & 0.549 & 0.513 & 0.607 & 0.521 & 0.543 & \textbf{0.754} \\
                & 50\%  & 0.553 & 0.431 & 0.555 & 0.494 & 0.793 & 0.682 & 0.711 & \textbf{0.924} \\
                & 100\% & 0.526 & 0.391 & 0.529 & 0.460 & 0.823 & 0.726 & 0.748 & \textbf{0.942} \\
                \midrule
                \multirow{3}{*}{Re-referencing}
                & Cz  & 0.517 & 0.435 & 0.520 & 0.597 & 0.950 & 0.937 & 0.987 & \textbf{0.998} \\
                & T7, T8  & 0.587 & 0.513 & 0.591 & 0.568 & 0.927 & 0.896 & 0.973 & \textbf{0.997} \\
                & bipolar & 0.752 & 0.621 & 0.757 & 0.719 & 0.940 & 0.919 & 0.962 & \textbf{0.996} \\
                \midrule
                \multirow{3}{*}{High-pass Filter}
                & 2 Hz  & 0.514 & 0.558 & 0.511 & 0.506 & 0.475 & 0.504 & 0.503 & \textbf{0.604} \\
                & 4 Hz  & 0.531 & 0.603 & 0.526 & 0.550 & 0.445 & 0.510 & 0.510 & \textbf{0.749} \\
                & 8 Hz & 0.446 & 0.684 & 0.439 & 0.637 & 0.367 & 0.546 & 0.544 & \textbf{0.872} \\
                \midrule
                \multirow{3}{*}{Low-pass Filter}
                & 45 Hz  & 0.499 & 0.502 & 0.499 & 0.503 & 0.461 & 0.507 & 0.505 & \textbf{0.822} \\
                & 30 Hz  & 0.500 & 0.511 & 0.498 & 0.515 & 0.246 & 0.649 & 0.644 & \textbf{0.986} \\
                & 15 Hz & 0.511 & 0.536 & 0.506 & 0.560 & 0.016 & 0.975 & 0.974 & \textbf{0.990} \\
                \bottomrule
            \end{tabular}
        \end{table*}

        OOD detection methods operating on discriminative models that were trained in a supervised way on the normality prediction task (MSP, Energy, ODIN, ASH) were largely ineffective for differentiating between in- and out-of-distribution data. These methods performed close to chance level (AUROC of 0.5) for most conditions and showed inconsistent trends with perturbation severity. Energy scores, for instance, showed increasingly better OOD detection performance with higher severities of the high-pass filter perturbation, but decreasing performance for increasing severities of the resampling perturbation.

        Generative models (in particular Typicality, DoSE, and SITN), on the other hand, achieved much better OOD detection overall with consistent severity-dependent performance improvements. SITN clearly emerged as the most effective OOD detection method with strong performance even at the lowest severity level and for perturbations where other methods performed close to chance level (see for example the channel shuffle perturbation at the lowest severity level where only two channels are flipped). Raw log-likelihoods worked well for certain perturbations (Channel Shuffle, Re-referencing), but not others, exhibiting a behaviour previously observed in the computer vision literature, which we discuss in more detail in \cref{sec:likelihood_limitations}.

        Evaluations on CAUEEG corroborate these results, except that DoSE narrowly outperformed SITN for the Re-referencing perturbation and the highest severity Low-pass Filter setting (see \cref{appendix:results_ood_detection_caueeg}). Additional results for Mahalanobis distances can be found in \cref{appendix:mahalanobis}.

    \subsection{Downstream Performance Impact}
    \label{sec:results_downstream_performance}

\begin{figure}[t]
            \centering
            \includegraphics[width=\textwidth]{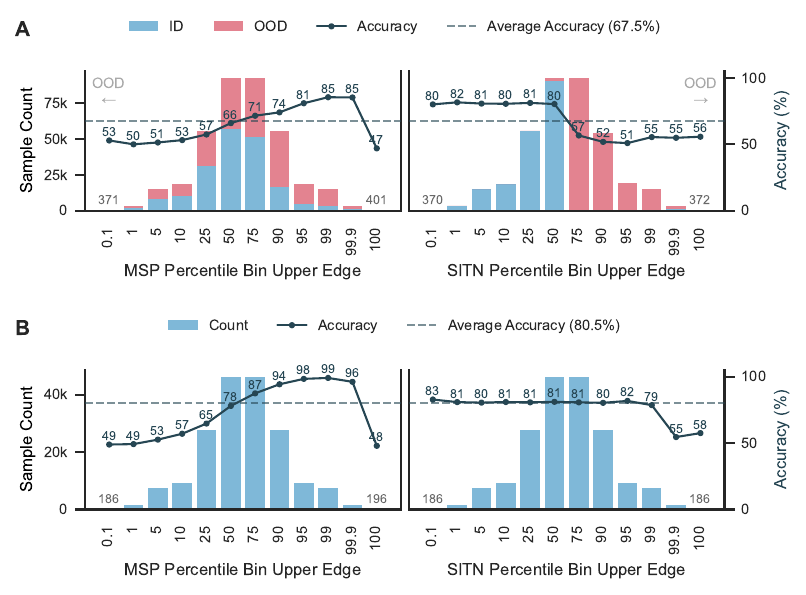}
            \caption{
                \textbf{Model performance for pathology prediction on the TUAB dataset.} \textbf{A} ID samples and corresponding OOD samples obtained with the resampling perturbation (severity 200 Hz). Performance is shown in terms of accuracy across percentile bins, where the x ticks indicate the upper edge of the corresponding bin (i.e. the first bin contains samples from the 0-0.1 percentile, the second one 0.1-1, etc.). The histograms indicate the number of ID and OOD samples in each bin. The arrows in the top left/right of each plot indicate the OOD direction for the corresponding metric. See \cref{appendix:full_downstream} for full downstream performance results across all perturbations and OOD detection methods.
                \textbf{B} The same visualisations restricted to only in-distribution data. MSP is a good indicator of model performance across in-distribution samples except for a failure mode at the highest MSP scores, whereas SITN provides little information about model performance between ID samples.
            }
            \label{fig:downsteam_performance}
        \end{figure}

        Panel A in \cref{fig:downsteam_performance} shows the impact of OOD detection on the clinical downstream EEG normality prediction task on the TUAB dataset (see \cref{sec:data}). The figure shows one method from each paradigm (MSP as a discriminative method and SITN as a generative method) where the resampling perturbation (severity 200 Hz) was used for the OOD data. For each method, the pooled in- and out-of-distribution samples were sorted according to the assigned OOD scores. Then we report the (downstream) classification accuracy of the discriminative model across different percentile bins of the OOD score (e.g. the performance across the samples with the lowest 0.1\% MSP scores, the 0.1-1\% lowest MSP scores, etc.). The arrows in the top left/right corners of the plots indicate the OOD direction for each method: for MSP, OOD samples are expected to have lower scores, whereas SITN should assign higher scores to OOD samples. Finally, the histograms show the sample counts in each percentile bin with the different colours indicating the proportion of ID and OOD samples.

        For MSP, the proportion of ID and OOD samples is relatively similar across the bins, consistent with the poor OOD detection AUROC values reported in \cref{sec:results_ood_detection}. MSP does not separate ID and OOD samples well.
        Despite this fact, MSP values are still predictive of model performance, except for a failure mode at very high confidence values.
        SITN, on the other hand, separates ID and OOD samples extremely well. It can be seen that the model performance is substantially lower on OOD data, dropping almost to chance level, such that OOD detection provides a very valuable safety net to prevent misclassifications.

        In Panel B, we present the same visualisations on ID data (i.e. the unperturbed test data) alone. It can be seen that MSP scores are much more predictive of model performance than SITN scores within ID samples. Generative approaches like SITN, which do not have access to the discriminative model used for the classification task, provide little indication for model performance within the majority of in-distribution samples, but they do not have the aforementioned failure mode of discriminative methods at very high confidence levels. A performance drop can be observed for ID samples with the most extreme (top 1\%) SITN scores. These samples likely represent rare in-distribution samples to which the model had very low exposure during training.

        Full results for all perturbations and OOD detection methods as well as evaluations on the dementia diagnosis task on CAUEEG can be found in  \cref{appendix:full_downstream}. While the effect on downstream performance was smaller for certain conditions (e.g. Low-pass Filter on CAUEEG; \cref{fig:appendix_downstream_lowpass_caueeg}), the general patterns are highly consistent.
        
        Overall, these results clearly disentangle two aspects that influence downstream performance. Generative methods like SITN excel at true OOD detection, identifying samples outside the training distribution, for which model predictions should not be trusted because the model was not exposed to comparable data during training. Discriminative methods like MSP, on the other hand, are not suitable for OOD detection, but reflect model uncertainty. Even with ample exposure to similar samples during training, certain cases can be ambiguous and difficult for the model to classify. Model uncertainty can capture such cases and provide valuable information about how well the model is likely to perform across different in-distribution samples.

    \subsection{Combination of OOD Detection and Model Uncertainty}
    
        The results in \cref{sec:results_downstream_performance} suggest that it could be extremely useful to combine OOD detection and model uncertainty methods. OOD detection methods can provide protection against highly confident mispredictions when the model encounters out-of-distribution samples, while model uncertainty methods function well for in-distribution data, where they provide valuable estimates of the model performance.

        \begin{figure}[t]
            \centering
            \includegraphics[width=\textwidth]{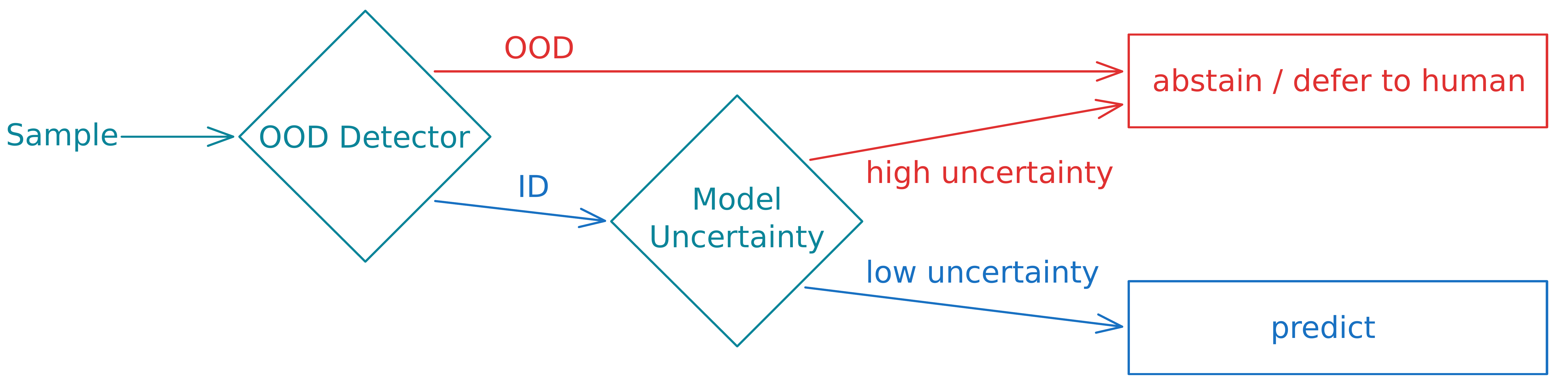}
            \caption{
                \textbf{Decision Diagram for Combined OOD and Uncertainty Detection.} Samples are jointly probed by an OOD detector and a model uncertainty estimator, each of which can trigger the system to abstain from prediction or defer to a human expert to prevent misclassifications.
            }
            \label{fig:decision_diagram}
        \end{figure}

        We evaluated a system following the decision diagram in \cref{fig:decision_diagram}, where samples are assessed by both an OOD detector and a model uncertainty estimator. The classifier then only provides predictions for samples that pass both tests and abstains from prediction otherwise. We used SITN for OOD detection, following the calibration procedure detailed in the original publication \cite{bomatter_signal_2026}, which allows one to pick a threshold with access to only an in-distribution validation set based on an acceptable false positive rate (FPR). A target FPR of 1\% was chosen for this example. To control model uncertainty, we first used isotonic regression (fitted on the in-distribution validation set) to map raw MSP values to calibrated probabilities. We then thresholded samples as too uncertain if the calibrated probability of the prediction being correct was below 60\%. The choice of this threshold will in practice depend on the costs of misclassifications vs abstaining from prediction or deferring to a human expert.

        Results on TUAB are shown in \cref{tab:results_ood_and_uncertainty}. Focusing on the baseline performance without any rejections (coverage 100\%) first, it can be seen that the accuracy dropped from 80.5\% on ID data to 67.5\% when corresponding OOD samples (resampling perturbation at 200 Hz) were included in the evaluation. When SITN was used to reject OOD samples, performance remained stable across both conditions. Note also that, as expected, the coverage on ID data is close to the desired 1\% FPR. Using MSP to reject high uncertainty samples improved performance on ID data (at the cost of a lower coverage). When OOD samples were included in the evaluation, MSP did not protect well against the drop in performance, with only marginally better accuracy than the baseline despite rejecting more than 10\% of the samples. Combining the two methods resulted in the best accuracy across both conditions.
        
        \begin{table}[h]
            \centering
            \caption{
                \textbf{Selective Classification Performance for OOD Detection and Model Uncertainty Estimation.} OOD Detection (using SITN), Model Uncertainty (using MSP), and their combination (SITN+MSP) are compared on ID-only data and ID+OOD data. OOD data was generated with the resampling perturbation at 200 Hz. MSP provides an accuracy improvement on in-distribution data, but only SITN provides effective protection against OOD data. Combining both methods enables the highest accuracy across both in- and out-of-distribution data.
            }
            \label{tab:results_ood_and_uncertainty}
            \begin{tabular}{@{}lcccc@{}}
            \toprule
             & \multicolumn{2}{c}{\textbf{ID Only}} & \multicolumn{2}{c}{\textbf{ID and OOD}} \\
            \cmidrule(lr){2-3} \cmidrule(l){4-5}
            \textbf{Rejection Method} & \textbf{Coverage} & \textbf{Accuracy} & \textbf{Coverage} & \textbf{Accuracy} \\
            \midrule
            Baseline (No Rejection) & 100.0\% & 80.5\% & 100.0\% & 67.5\% \\
            SITN (OOD Detection) & 98.5\%  & 80.8\% & 49.5\%  & 80.7\% \\
            MSP (Model Uncertainty)  & 88.1\%  & 83.8\% & 89.2\%  & 69.3\% \\
            Combined (SITN + MSP)   & 86.9\%  & 84.1\% & 43.7\%  & 84.1\% \\
            \bottomrule
            \end{tabular}
        \end{table}

\section{Discussion and Conclusions}
\label{sec:discussion}

    In this study, we introduced an OOD detection benchmark for EEG-based machine learning and systematically evaluated a broad spectrum of both discriminative and generative methods. Our results clearly demonstrate that standard discriminative methods are incapable of reliably detecting OOD EEG data under realistic covariate shifts, a finding that aligns with limited prior work in the EEG BCI literature \cite{mulder_challenge_2026, liu_temporal_2026}. However, generative approaches---which to the best of our knowledge have not been previously benchmarked for EEG or general time-series OOD detection---yield substantially better performance. In particular, Signal in the Noise (SITN) establishes a new state of the art for EEG OOD detection.

    Follow-up evaluations to assess the utility of OOD detection in two clinical downstream prediction tasks demonstrated two things. First, performance dropped substantially on out-of-distribution samples, emphasising the value of OOD detection to prevent mispredictions in high-risk applications. 
    Second, both discriminative and generative methods were predictive of downstream performance, but for different reasons. Only generative methods protect against model failure by flagging OOD samples. Discriminative methods prevent mispredictions by highlighting samples with high model uncertainty, even if they are in-distribution. These results provide strong support for the position recently put forward by \citeauthor{li_position_2025}, which argues that discriminative methods "answer the wrong question for OOD detection"\cite{li_position_2025}. We furthermore demonstrated how the complementary capabilities of discriminative and generative methods can be combined to protect against both failure modes.

    The present study has some limitations that highlight valuable directions for future work. While the creation of OOD data through the perturbations presented in this work has several advantages (unambiguous ground truth, perfect pairing with ID samples, large sample size, control of severity levels), the field would benefit from additional benchmarks and curated datasets to study a wider variety of distribution shifts.
    Designing such OOD detection setups is more challenging than in computer vision. While images from different datasets (e.g. CIFAR-10 vs SVHN) or different object categories (e.g. cats vs cars) are clearly distinct, comparing EEG segments across different datasets is more nuanced. Besides clear differences in recording hardware or preprocessing, aspects of which are already reflected in our perturbation experiments, the degree to which differences in patient populations manifest in EEG distribution shifts is hard to quantify. Even for comparatively well-studied variables like sex or age with known population-level differences \cite{aurlien_eeg_2004, chiang_age_2011, cellier_development_2021}, a lot of the variance is driven by individual differences. Moreover, cross-dataset label definitions are often not mutually exclusive. EEG segments in the TUAB dataset could in principle also be assigned a simultaneous label for the dementia diagnosis task in CAUEEG, such that CAUEEG segments need not necessarily be OOD for a TUAB-trained model (see exploratory analysis in \cref{appendix:cross_dataset}). Where possible, we recommend grounding OOD detection through simultaneous assessments of performance in a relevant downstream task as in our experiments. This ensures that progress in OOD detection will translate to superior safeguards in practical applications.

    Our results also raise questions about benchmarking practices in the broader OOD detection literature. How can the poor performance of discriminative methods in our evaluations be reconciled with their success on widely used computer vision benchmarks like OpenOOD?
    We hypothesise that this discrepancy stems from fundamental differences in how OOD evaluation is framed. For one, established leaderboards frequently permit the use of auxiliary OOD data during training. Our setup, by contrast, strictly assumes no prior access to OOD data, reflecting a more realistic scenario where models encounter completely unanticipated shifts.
    Furthermore, OOD detection benchmarks are dominated by semantic shifts \cite{guille-escuret_expecting_2024}. An influential survey paper by \citeauthor{yang_generalized_2024} even frames this exclusive focus as part of the general definition of OOD detection, although it is acknowledged that other shifts may be considered OOD from a generalisation perspective or when considering high-risk applications \cite{yang_generalized_2024}.
    We encourage future applied work to place practical utility at its centre. In clinical domains, the primary objective of OOD detection is to prevent model failure caused by unanticipated distribution shifts. Crucially, achieving this requires benchmarks designed to prevent the entanglement of true OOD detection and model uncertainty. As our results suggest, these are fundamentally distinct capabilities and conflating the two obscures what a given method actually achieves. By designing benchmarks that explicitly separate OOD detection from model uncertainty estimation, the field can make targeted progress on both fronts.

\newpage
\backmatter

\bmhead{Acknowledgements}

    This project was supported by the Royal Academy of Engineering under the Research Fellowship programme.

\bibliography{bibliography}

\newpage
\begin{appendices}
\crefalias{section}{appendix}

    \section{OOD Detection Results on CAUEEG}
    \label{appendix:results_ood_detection_caueeg}

        \begin{table*}[h]
            \centering
            \small
            \setlength{\tabcolsep}{3pt}
            \caption{
                \textbf{OOD detection performance in terms of AUROC across different perturbations and severity levels on CAUEEG.}
                Bold values indicate the best performing metric for a given perturbation and severity.
                Discriminative-model-based methods (MSP, Energy, ODIN, ASH) perform largely at chance level, whereas generative-model-based methods achieve considerably better OOD detection with increasing performance for higher severity levels.
            }
            \label{tab:ood_auroc_caueeg}
            \begin{tabular}{ll cccccccc}
                \toprule
                \textbf{Perturbation} & \textbf{Severity} & \textbf{MSP} & \textbf{Energy} & \textbf{ODIN} & \textbf{ASH} & \textbf{LL} & \textbf{Typ.} & \textbf{DoSE} & \textbf{SITN} \\
                \midrule
                \multirow{3}{*}{Resampling}
                & 125 Hz & 0.531 & 0.517 & 0.540 & 0.540 & 0.286 & 0.587 & 0.760 & \textbf{0.956} \\
                & 150 Hz & 0.497 & 0.520 & 0.514 & 0.590 & 0.166 & 0.732 & 0.886 & \textbf{0.984} \\
                & 200 Hz & 0.470 & 0.531 & 0.495 & 0.658 & 0.056 & 0.906 & 0.965 & \textbf{0.992} \\
                \midrule
                \multirow{3}{*}{Channel Shuffle}
                & 10\%  & 0.496 & 0.502 & 0.505 & 0.518 & 0.622 & 0.552 & 0.711 & \textbf{0.759} \\
                & 50\%  & 0.443 & 0.490 & 0.470 & 0.570 & 0.841 & 0.766 & 0.968 & \textbf{0.956} \\
                & 100\% & 0.408 & 0.483 & 0.436 & 0.583 & 0.879 & 0.819 & 0.982 & \textbf{0.968} \\
                \midrule
                \multirow{3}{*}{Re-referencing}
                & Cz  & 0.281 & 0.230 & 0.305 & 0.366 & 0.992 & 0.991 & \textbf{1.000} & 0.997 \\
                & T7, T8  & 0.305 & 0.361 & 0.339 & 0.471 & 0.982 & 0.976 & \textbf{1.000} & 0.997 \\
                & bipolar & 0.426 & 0.439 & 0.454 & 0.562 & 0.990 & 0.989 & \textbf{1.000} & 0.997 \\
                \midrule
                \multirow{3}{*}{High-pass Filter}
                & 2 Hz  & 0.520 & 0.640 & 0.512 & 0.532 & 0.455 & 0.507 & 0.566 & \textbf{0.606} \\
                & 4 Hz  & 0.507 & 0.685 & 0.494 & 0.578 & 0.408 & 0.518 & 0.594 & \textbf{0.756} \\
                & 8 Hz & 0.518 & 0.792 & 0.494 & 0.721 & 0.291 & 0.589 & 0.626 & \textbf{0.899} \\
                \midrule
                \multirow{3}{*}{Low-pass Filter}
                & 45 Hz  & 0.496 & 0.501 & 0.496 & 0.502 & 0.446 & 0.505 & 0.544 & \textbf{0.822} \\
                & 30 Hz  & 0.495 & 0.510 & 0.492 & 0.515 & 0.186 & 0.700 & 0.943 & \textbf{0.987} \\
                & 15 Hz & 0.557 & 0.564 & 0.556 & 0.563 & 0.007 & 0.989 & \textbf{0.998} & 0.994 \\
                \bottomrule
            \end{tabular}
        \end{table*}

    \section{Mahalanobis Distance}
    \label{appendix:mahalanobis}

        We also evaluated the Mahalanobis distance \cite{lee_simple_2018}, an OOD detection approach that falls into the discriminative category but differs from the methods detailed in \cref{sec:ood_methods} by modelling the density of the latent embedding space, rather than deriving a score based on the network's predictions.
        
        Specifically, we first extracted the penultimate-layer feature representations from the pre-trained discriminative model, taking the activations immediately following the global average pooling layer. Using the in-distribution validation set, we estimated the class-conditional mean feature vectors alongside a single, shared covariance matrix tied across all classes. To ensure numerical stability, a small regularisation term ($\epsilon = 10^{-5}$) was added to the diagonal of the shared covariance matrix prior to computing its inverse (the shared precision matrix). During evaluation, the Mahalanobis distance was calculated between each test sample's feature vector and each of the class-conditional means using this shared precision matrix. A sample's final score was then defined as its minimum Mahalanobis distance across all available classes. OOD samples are expected to have higher scores indicating that their embeddings are situated further from the known in-distribution classes.

        OOD detection results are shown in \cref{tab:ood_auroc_mahalanobis}. In contrast to other discriminative methods, the Mahalanobis distance yielded superior OOD detection performance under the Resampling, Channel Shuffle, and Re-referencing perturbations. However, it exhibited a severe vulnerability to the High-pass Filter perturbation, where it performed worse than random chance. 

        \begin{table*}[h]
            \centering
            \small
            \setlength{\tabcolsep}{3pt}
            \caption{
                \textbf{Mahalanobis distance OOD detection performance in terms of AUROC across different perturbations and severity levels.} Compared to other discriminative approaches, Mahalanobis distances show better performance for the Resampling, Channel Shuffle, and Re-referencing perturbations, but also a strong failure for the High-pass Filter perturbation, where they consistently underperform even random guessing.
            }
            \label{tab:ood_auroc_mahalanobis}
            \begin{tabular}{ll cc}
                \toprule
                \textbf{Perturbation} & \textbf{Severity} & \textbf{TUAB} & \textbf{CAUEEG} \\
                \midrule
                \multirow{3}{*}{Resampling}
                & 125 Hz  & 0.548 & 0.544 \\
                & 150 Hz  & 0.594 & 0.586 \\
                & 200 Hz  & 0.654 & 0.640 \\
                \midrule
                \multirow{3}{*}{Channel Shuffle}
                & 10\%    & 0.583 & 0.604 \\
                & 50\%    & 0.741 & 0.763 \\
                & 100\%   & 0.775 & 0.799 \\
                \midrule
                \multirow{3}{*}{Re-referencing}
                & Cz      & 0.851 & 0.925 \\
                & T7, T8  & 0.794 & 0.845 \\
                & bipolar & 0.786 & 0.861 \\
                \midrule
                \multirow{3}{*}{High-pass Filter}
                & 2 Hz    & 0.348 & 0.263 \\
                & 4 Hz    & 0.321 & 0.252 \\
                & 8 Hz    & 0.285 & 0.153 \\
                \midrule
                \multirow{3}{*}{Low-pass Filter}
                & 45 Hz   & 0.500 & 0.501 \\
                & 30 Hz   & 0.506 & 0.505 \\
                & 15 Hz   & 0.541 & 0.519 \\
                \bottomrule
            \end{tabular}
        \end{table*}

    \section{Sample Size Dependence}

        A potential limitation of generative methods could be a larger sample size requirement. To investigate this possibility, we performed a data scaling analysis where the amount of training data was varied while the validation and test data was kept the same. Starting from a subset of just 25 participants and 5 segments per participant, we increased the sample size in two different ways: either increasing the number of participants or the number of segments per participant. These experiments were conducted on the TUAB dataset and subsampling of participants was stratified by the normality label. We used the resampling perturbation at 200 Hz for OOD detection evaluations.

\begin{figure}[H]
            \centering
            \includegraphics[width=\textwidth]{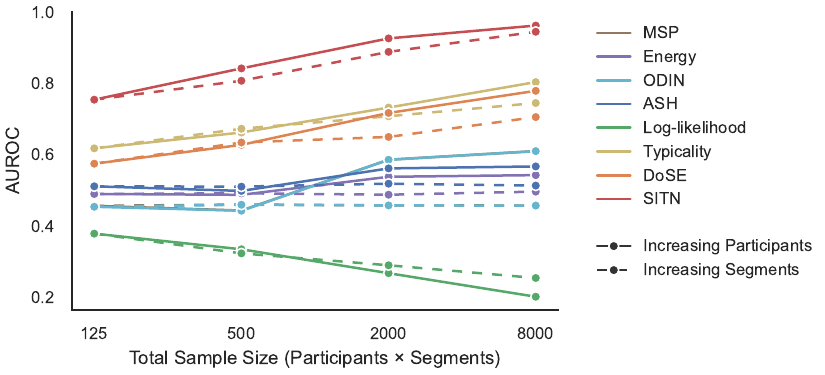}
            \caption{
                \textbf{Sample Size Dependence of OOD Detection Performance.} Starting from a baseline of just 25 participants and 5 segments per participant (= 125 samples in total), the training data was gradually increased by either increasing the number of participants (solid lines) or segments per participant (dashed lines). The sampling rate perturbation at 200 Hz was used to generate OOD samples. OOD detection performance generally increased with the amount of training data (see \cref{sec:likelihood_limitations} for a discussion of the likelihood). Even in the smallest data regimes, generative methods dominated discriminative approaches. Note that MSP is occluded by ODIN, which showed almost identical performance across sample sizes.
            }
            \label{fig:sample_size}
        \end{figure}

        As shown in \cref{fig:sample_size}, OOD detection performance generally increased with the amount of training data, where a larger number of participants (solid lines) was more impactful than an equal number of segments from a smaller participant cohort (dashed lines). Generative methods (SITN in particular) performed surprisingly well even with very limited data and dominated discriminative methods across all evaluated sample sizes. The failure mode for log-likelihoods where performance decreased with more training data is discussed in \cref{sec:likelihood_limitations}.

    \section{Limitations of Likelihoods for OOD Detection}
    \label{sec:likelihood_limitations}

        Several works in the computer vision literature have identified critical limitations of likelihoods for OOD detection \cite{nalisnick_deep_2019, serra_input_2020, schirrmeister_understanding_2020, kirichenko_why_2020, havtorn_hierarchical_2022, kamkari_geometric_2024}, which motivated the development of methods like Typicality, DoSE, and SITN. In widely replicated experiments, it was shown that generative models trained on CIFAR-10 consistently assign higher likelihoods to samples from other datasets like MNIST or SVHN \cite{nalisnick_deep_2019, serra_input_2020, schirrmeister_understanding_2020, kirichenko_why_2020}. Investigating this issue, \citeauthor*{serra_input_2020} further observed a strong correlation between likelihood and sample complexity (measured by PNG compression size), showing that "simpler" samples like constant-colour images were consistently assigned higher likelihoods than more "complex" ones.

\begin{figure}[H]
            \centering
            \includegraphics[width=\textwidth]{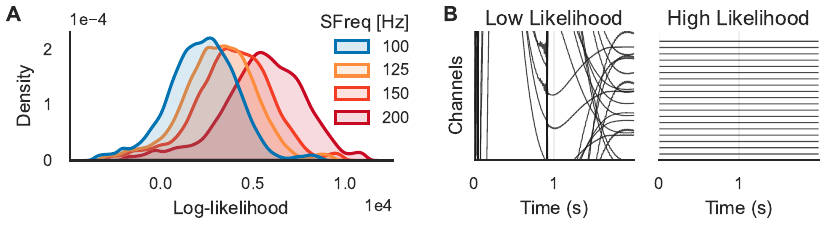}
            \caption{
                \textbf{Likelihood Limitations.}
                \textbf{A} Distribution of log-likelihoods across ID samples (100 Hz) and OOD samples obtained with the resampling perturbation at increasing severities of 125, 150, and 200 Hz. As previously observed in the context of computer vision datasets, OOD samples are consistently assigned higher likelihoods than the in-distribution data.
                \textbf{B} EEG segments that were assigned the lowest and highest likelihoods, respectively. As in computer vision, where "simpler" images like constant-colour images are assigned high likelihoods, EEG segments with completely flat channels also result in high likelihoods. High-amplitude artefacts result in the opposite extreme and are assigned very small likelihoods.
            }
            \label{fig:likelihood_failures}
        \end{figure}

        We investigated and were able to replicate the same behaviour for EEG data. Panel A in \cref{fig:likelihood_failures} shows the log-likelihood distribution across ID test samples (100 Hz) compared to the distributions for the corresponding OOD samples obtained through the resampling perturbation at different severities. It can be seen that OOD samples were consistently assigned higher likelihoods, with the highest likelihoods assigned to samples perturbed at the highest severity level. As mentioned in \cref{sec:perturbation_suite} and illustrated in \cref{fig:perturbations}, samples perturbed with this perturbation appear increasingly smoothed with higher severities. Furthermore, Panel B in the same figure shows the (ID) EEG segments with the lowest and highest likelihood in the dataset. Consistent with the complexity observations by \citeauthor{serra_input_2020}, the highest likelihoods are assigned to completely flat EEG segments, whereas high-amplitude artefacts resulted in particularly low likelihoods.

    \section{Training Details}
    \label{appendix:training_details}

        For the generative model, the UNet architecture following \citeauthor*{ho_denoising_2020} \cite{ho_denoising_2020}, as implemented in \citeauthor*{dhariwal_diffusion_2021} \cite{dhariwal_diffusion_2021} was used with the hyperparameters detailed in \cref{tab:hyperparameters_unet}.
        Following the flow matching objective of \citeauthor*{lipman_flow_2023} \cite{lipman_flow_2023}, we constructed a linear interpolant between a Gaussian noise sample $Z \sim \mathcal{N}(0,I)$ and a data sample $X \sim P_{data}$: $X_t = (1-t)Z + tX, \quad t \sim \mathcal{U}[0,1]$ and minimised the mean-squared error between the network output and the target vector field $X - Z$.
        
        For the discriminative model, a Temporal Convolutional Network (TCN) \cite{bai_empirical_2018} was used with the hyperparameters detailed in \cref{tab:hyperparameters_tcn}.
        This model was trained using a supervised cross-entropy loss for the respective prediction task.
        
        Both models were optimised with AdamW (learning rate $10^{-4}$, weight decay $0.01$, $\beta_1=0.9$, $\beta_2=0.999$, $\epsilon=10^{-8}$) and global gradient norm clipping (max norm $1.0$).
        The generative and discriminative models were trained for 1{,}000{,}000 and 50{,}000 gradient steps, respectively, at a batch size of 128 and with weights restored to the checkpoint of lowest validation loss. It was empirically verified that the validation loss had converged within the maximum number of gradient steps for both models.

        \begin{table}[h]
            \makebox[\textwidth][c]{%
                \hspace{2cm}
                \begin{minipage}[t]{0.5\textwidth}
                    \centering
                    \caption{UNet hyperparameters.}
                    \label{tab:hyperparameters_unet}
                    \begin{tabular}{ll}
                        \toprule
                        Hyperparameter & Value \\
                        \midrule
                        Channels              & 64 \\
                        Depth                 & 2 \\
                        Channel multipliers   & 1, 2, 2, 2 \\
                        Heads                 & 1 \\
                        Attention resolution  & 8 \\
                        Dropout               & 0.0 \\
                        \bottomrule
                    \end{tabular}
                \end{minipage}%
                \hspace{1cm}
                \begin{minipage}[t]{0.5\textwidth}
                    \centering
                    \caption{TCN hyperparameters.}
                    \label{tab:hyperparameters_tcn}
                    \begin{tabular}{ll}
                        \toprule
                        Hyperparameter & Value \\
                        \midrule
                        Blocks                & 4 \\
                        Filters               & 64 \\
                        Kernel size           & 5 \\
                        Dilations             & 1, 2, 4, 8 \\
                        Dropout               & 0.0 \\
                        \bottomrule
                    \end{tabular}
                \end{minipage}%
            }
        \end{table}

    \section{Exploratory Cross-dataset Experiments}
    \label{appendix:cross_dataset}

        We conducted a number of cross-dataset experiments where we evaluated the TUAB-trained methods on CAUEEG and conversely the CAUEEG-trained methods on TUAB. We consider this analysis exploratory as it is unclear to what degree samples from one of these datasets should be considered OOD for a model trained on the other dataset. Aside from potential (unquantified) hardware differences, participants in the TUAB dataset could in principle also be assigned a label from the dementia diagnosis task on CAUEEG, such that EEG recordings from one dataset should not necessarily be considered OOD for a model trained on the other one. Rather than treating this setup as a benchmarking task to evaluate OOD detection performance, we are interested to what degree cross-dataset samples result in high OOD scores and if there are differences across participant subgroups.

        \cref{tab:cross_dataset_auroc} shows AUROC values analogous to those reported for OOD detection performance. Overall, the different methods show moderate discrimination between the two datasets. Scores from DoSE and SITN, the two strongest OOD detection methods according to the results in \cref{sec:results_ood_detection}, do not separate samples from the two datasets when trained on TUAB (AUROC $<$ 0.55), but achieve AUROC values of 0.703 and 0.740 respectively when trained on CAUEEG. This would be consistent with CAUEEG forming a subset of the TUAB data distribution.

        \begin{table*}[h]
            \centering
            \small
            \setlength{\tabcolsep}{3pt}
            \caption{
                \textbf{Cross-dataset Discrimination Across OOD Detection Methods.}
                Models were trained on one dataset and then evaluated using the test split of the training dataset as ID and the test split of the other dataset as OOD data (e.g. TUAB$\rightarrow$CAUEEG denotes trained on TUAB and using CAUEEG as OOD). The degree to which OOD scores discriminate between samples from the two datasets is reported in terms of AUROC. Note that ID and OOD should be interpreted with a grain of salt in this setup as it is unclear to what degree cross-dataset samples are OOD. 
            }
            \label{tab:cross_dataset_auroc}
            \begin{tabular}{l cccccccc}
                \toprule
                \textbf{Condition} & \textbf{MSP} & \textbf{Energy} & \textbf{ODIN} & \textbf{ASH} & \textbf{LL} & \textbf{Typ.} & \textbf{DoSE} & \textbf{SITN} \\
                \midrule
                TUAB $\rightarrow$ CAUEEG & 0.594 & 0.478 & 0.598 & 0.517 & \textbf{0.669} & 0.432 & 0.462 & 0.544 \\
                CAUEEG $\rightarrow$ TUAB & 0.489 & 0.558 & 0.477 & 0.514 & 0.398 & 0.643 & 0.703 & \textbf{0.740} \\
                \bottomrule
            \end{tabular}
        \end{table*}

        In the following figures, we compare OOD scores across different subgroups within the target dataset. To do so, we aggregated OOD scores on the participant level by averaging across samples, resulting in an average OOD score per participant. Furthermore, to facilitate comparisons, we normalised OOD scores by applying min-max scaling to map the raw outputs of each OOD detection method to a standard $[0, 1]$ range. Additionally, scores were inverted where necessary, such that a higher normalised OOD score always indicates "more OOD". 

\begin{figure}[H]
            \centering
            \includegraphics[width=\textwidth]{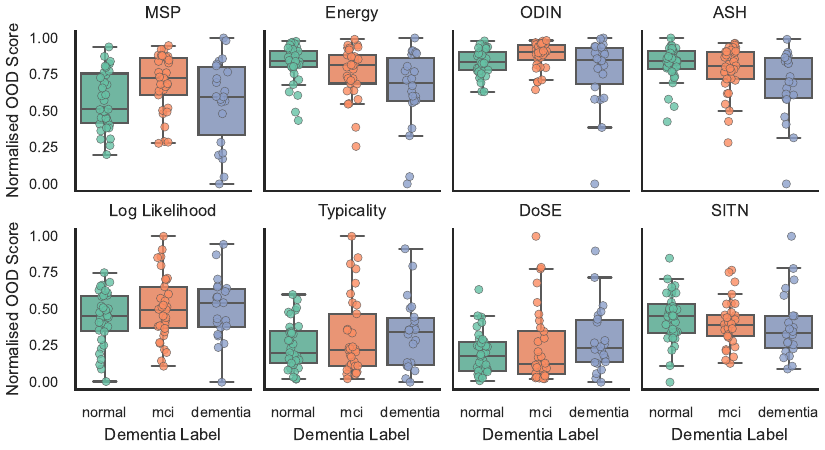}
            \caption{
                \textbf{Cross-dataset Subgroup Analysis of OOD Scores Across Diagnosis Labels on CAUEEG.} The results show normalised participant-level OOD scores for samples from the CAUEEG dataset assigned by methods trained on TUAB. Scores are largely overlapping without a consistent trend across diagnosis groups.
            }
            \label{fig:cross_dataset_dementia_labels}
        \end{figure}

\begin{figure}[H]
            \centering
            \includegraphics[width=\textwidth]{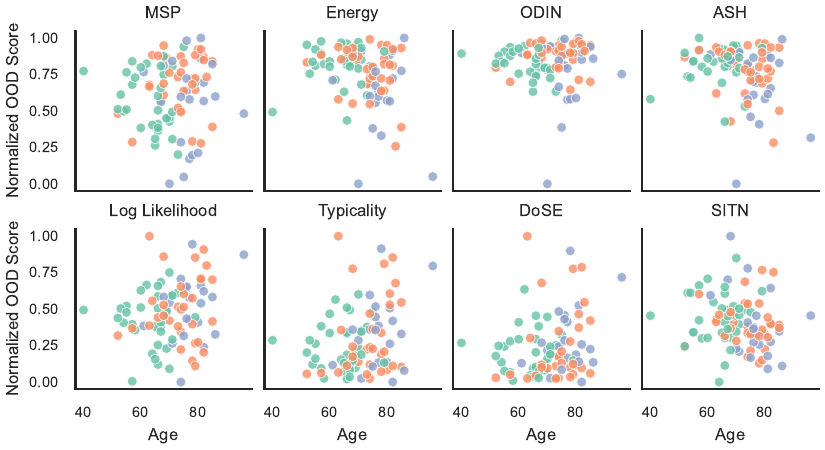}
            \caption{
                \textbf{Cross-dataset Subgroup Analysis of OOD Scores Across Age on CAUEEG.} The results show normalised participant-level OOD scores for samples from the CAUEEG dataset assigned by methods trained on TUAB. No clear trend is observable.
            }
            \label{fig:cross_dataset_age_on_caueeg}
        \end{figure}

\begin{figure}[H]
            \centering
            \includegraphics[width=\textwidth]{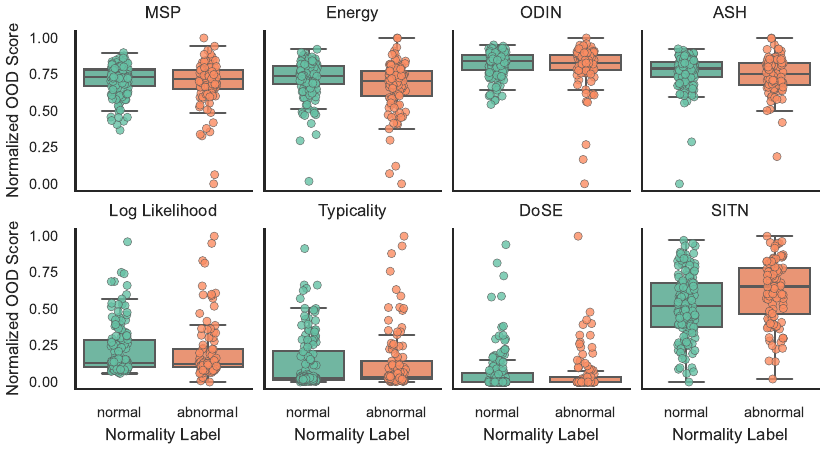}
            \caption{
                \textbf{Cross-dataset Subgroup Analysis of OOD Scores Across EEG Normality Labels on TUAB.} The results show normalised participant-level OOD scores for samples from the TUAB dataset assigned by methods trained on CAUEEG. Scores are largely overlapping, although SITN assigned somewhat higher OOD scores to abnormal EEG recordings.
            }
            \label{fig:cross_dataset_normality_labels}
        \end{figure}

\begin{figure}[H]
            \centering
            \includegraphics[width=\textwidth]{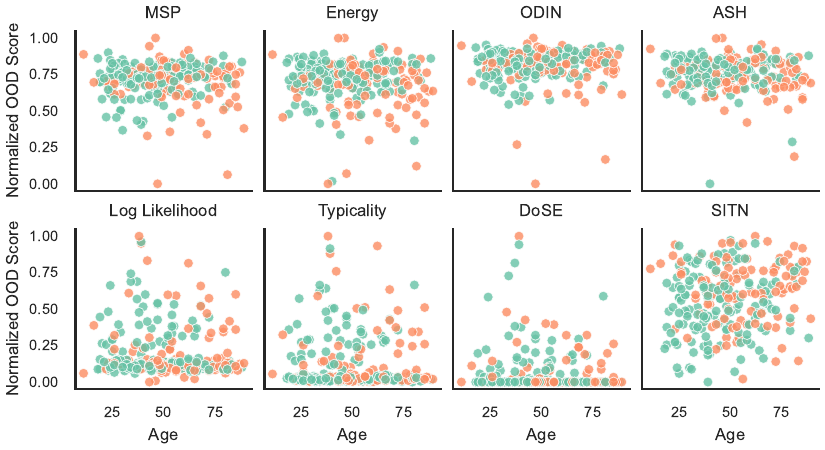}
            \caption{
                \textbf{Cross-dataset Subgroup Analysis of OOD Scores Across Age on TUAB.} The results show normalised participant-level OOD scores for samples from the TUAB dataset assigned by methods trained on CAUEEG. No clear trends are discernable.
            }
            \label{fig:cross_dataset_age_on_tuab}
        \end{figure}

        \cref{fig:cross_dataset_dementia_labels} shows normalised OOD scores across the different diagnosis labels on CAUEEG (normal, mild cognitive impairment, dementia) for the TUAB-trained models. Trends across diagnosis groups are inconsistent across methods and scores are strongly overlapping. \cref{fig:cross_dataset_age_on_caueeg} shows the same scores across age, with the colours corresponding to the same diagnosis groups (normal in green, mci in orange, dementia in blue). No clear trends are discernable, which might be expected given that the age range of the TUAB training data spans the full age range of the CAUEEG dataset. \cref{fig:cross_dataset_normality_labels} shows normalised scores for the CAUEEG-trained models on TUAB, stratified by the EEG normality label. Although scores are strongly overlapping for all methods, abnormal EEGs tended to be assigned somewhat higher OOD scores by SITN. Finally, \cref{fig:cross_dataset_age_on_tuab} shows the same OOD scores across age on TUAB. As for CAUEEG, no clear trend is discernable, although in this case one might have expected younger participants to look more OOD given that only 14 out of 950 participants in the CAUEEG training data (less than 2\%) are aged younger than 50.
        
    \section{Cross-participant BCI Experiment}

        We evaluated the different OOD detection methods in a cross-participant setup on the BNCI 2014-001 Motor Imagery dataset \cite{tangermann_review_2012}. This dataset contains EEG data for a motor imagery task from nine participants. Two sessions (recorded on different days) are available for each participant, each of which in turn consists of six runs with 12 trials per class. In our setup, we only used trials for the left hand and right hand classes, resulting in 144 trials per session and participant. We used the official train-test split, which assigns the first session of each participant to train and the second to test. Additionally, we used the last run in the training session as validation data. For our cross-participant experiment, we trained the classification model and the OOD detection methods only on the training data of the first participant. We then evaluated the methods on the test data of the training participant as well as on the test data of all other participants.

\begin{figure}[H]
            \centering
            \includegraphics[width=\textwidth]{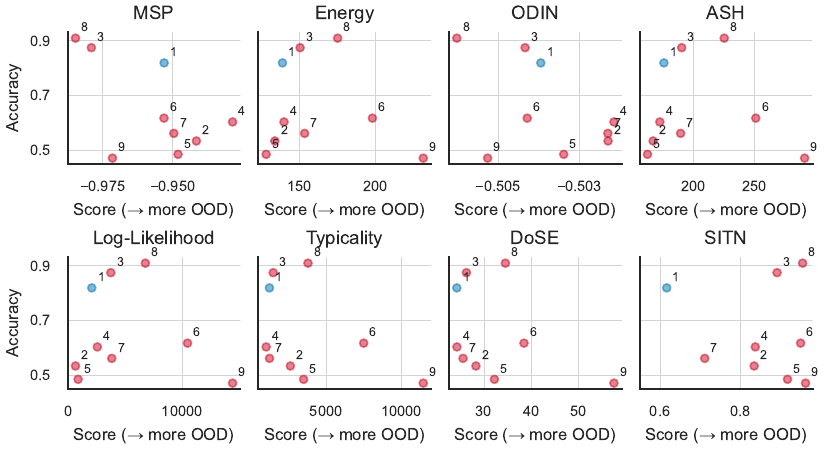}
            \caption{
                \textbf{Cross-participant Experiments on the BNCI 2014-001 Motor Imagery Dataset.} 
                Numbers next to the points indicate participant id. Methods were trained on the training data from participant 1 and evaluated on the test data from all participants. OOD scores were aggregated on the participant-level through averaging and aligned such that higher scores always indicate more OOD. Only SITN clearly separates test data from the training participant (ID) from the other participants (OOD). MSP assigns the lowest OOD scores to two participants where model performance was even higher than for the training participant, but also the 3rd lowest score to the participant where model performance was lowest and which was flagged as highly OOD by most other methods. 
            }
            \label{fig:cross_participants}
        \end{figure}

        \cref{fig:cross_participants} shows the results with prediction accuracy for the motor imagery task and OOD scores aggregated to the participant level through averaging. For better visualisation, we flipped scores where necessary, such that higher scores on the x-axis always correspond to higher OOD scores. SITN is the only method to clearly separate the (ID) training participant from all (OOD) participants. On six out of eight of the unseen participants prediction accuracy for the motor imagery task dropped considerably below the within-participant test performance. MSP assigned the lowest OOD scores to the two participants where model performance was higher than for the training participant, however, it also assigned the third lowest OOD score to a participant where performance dropped below chance level and which was flagged as highly OOD by all generative methods as well as the Energy score and ASH.

    \section{Full Downstream Performance Results}
    \label{appendix:full_downstream}

        The following pages show the full results for the downstream performance evaluations across both tasks and datasets (normality prediction on TUAB (binary classification, chance level 50\%); dementia diagnosis on CAUEEG (3-way classification into normal, MCI, dementia, chance level $\approx$ 33\%)) for all perturbations and OOD detection methods. Performance is shown in terms of accuracy across percentile bins of the OOD score, where the x ticks indicate the upper edge of the corresponding bin (i.e. the first bin contains samples from the 0-0.1 percentile, the second one 0.1-1, etc.). The histograms indicate the number of ID and OOD samples in each bin. The arrows in the top left/right of each plot indicate the OOD direction for the corresponding metric. The captions for each figure indicate the task and dataset as well as the perturbation used to generate the OOD data.

        \newpage
        \begin{figure}[H]
            \centering
            \includegraphics[width=\textwidth]{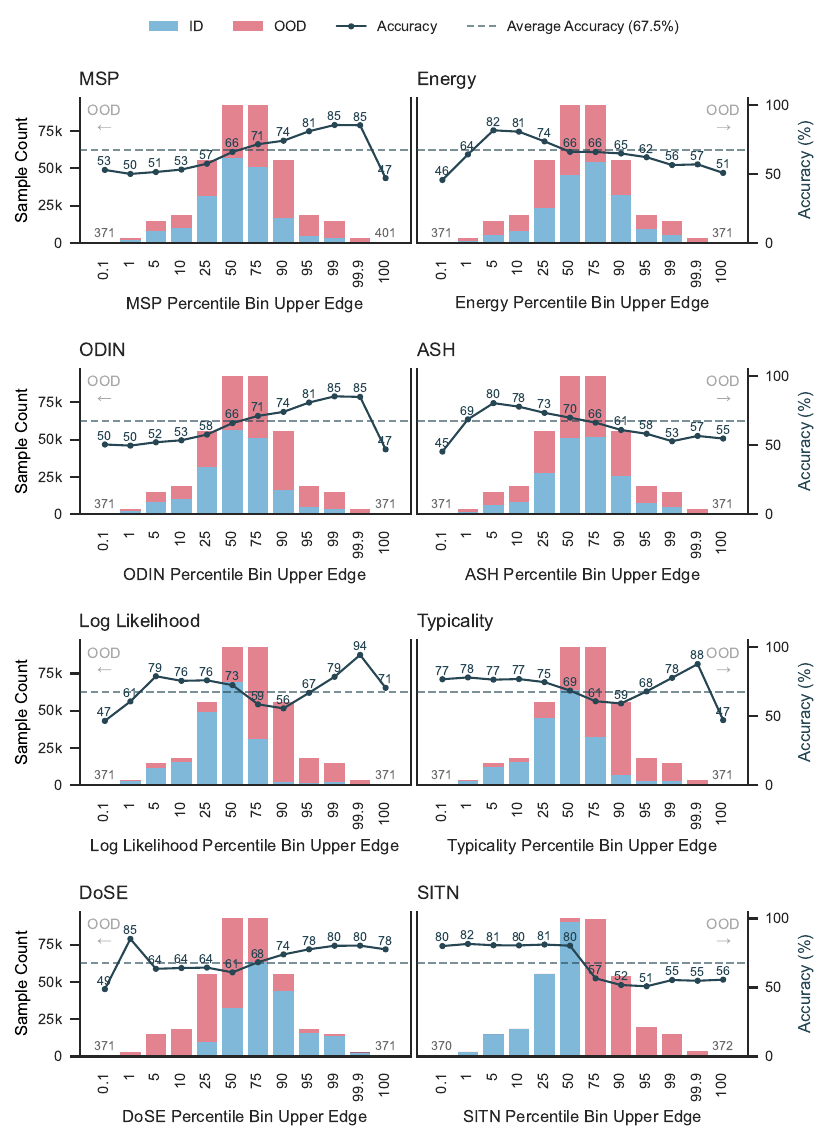}
            \caption{
                \textbf{Normality Prediction on TUAB --- Resampling Perturbation (200 Hz)}
            }
        \end{figure}

        \begin{figure}[H]
            \centering
            \includegraphics[width=\textwidth]{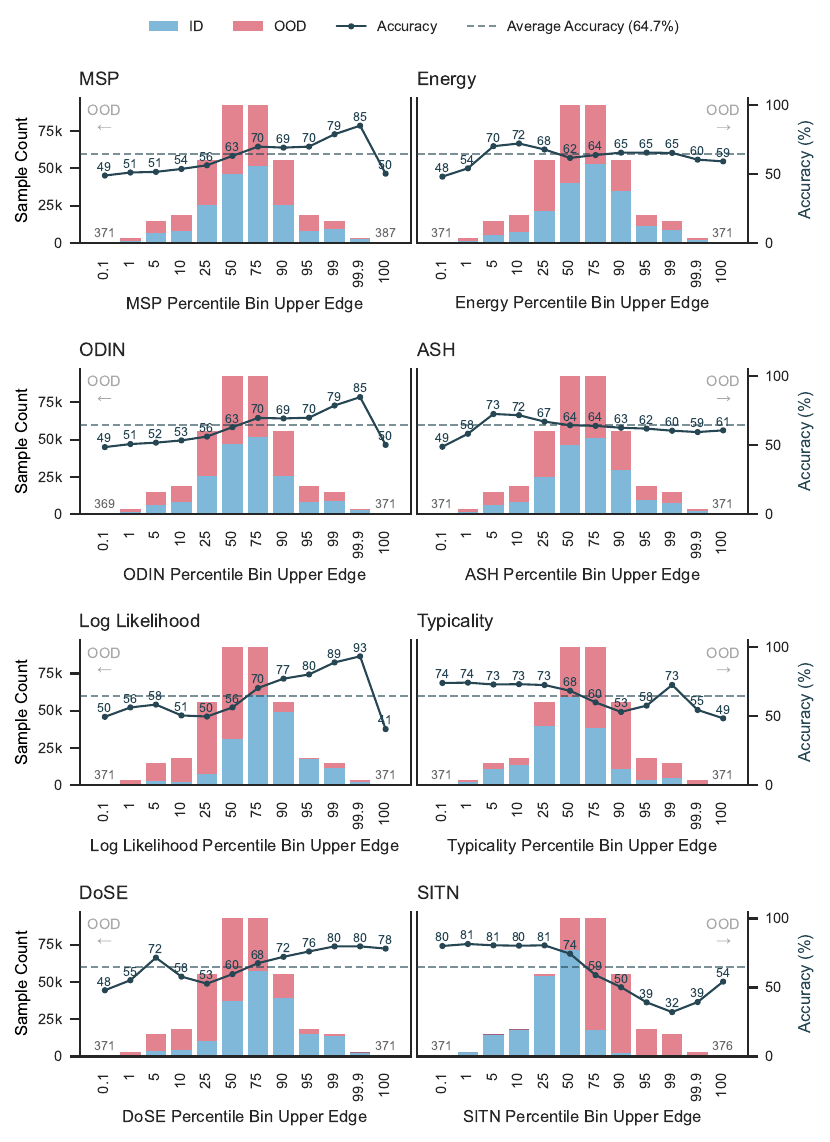}
            \caption{
                \textbf{Normality Prediction on TUAB --- Channel Shuffle Perturbation (100\%)}
            }
        \end{figure}

        \begin{figure}[H]
            \centering
            \includegraphics[width=\textwidth]{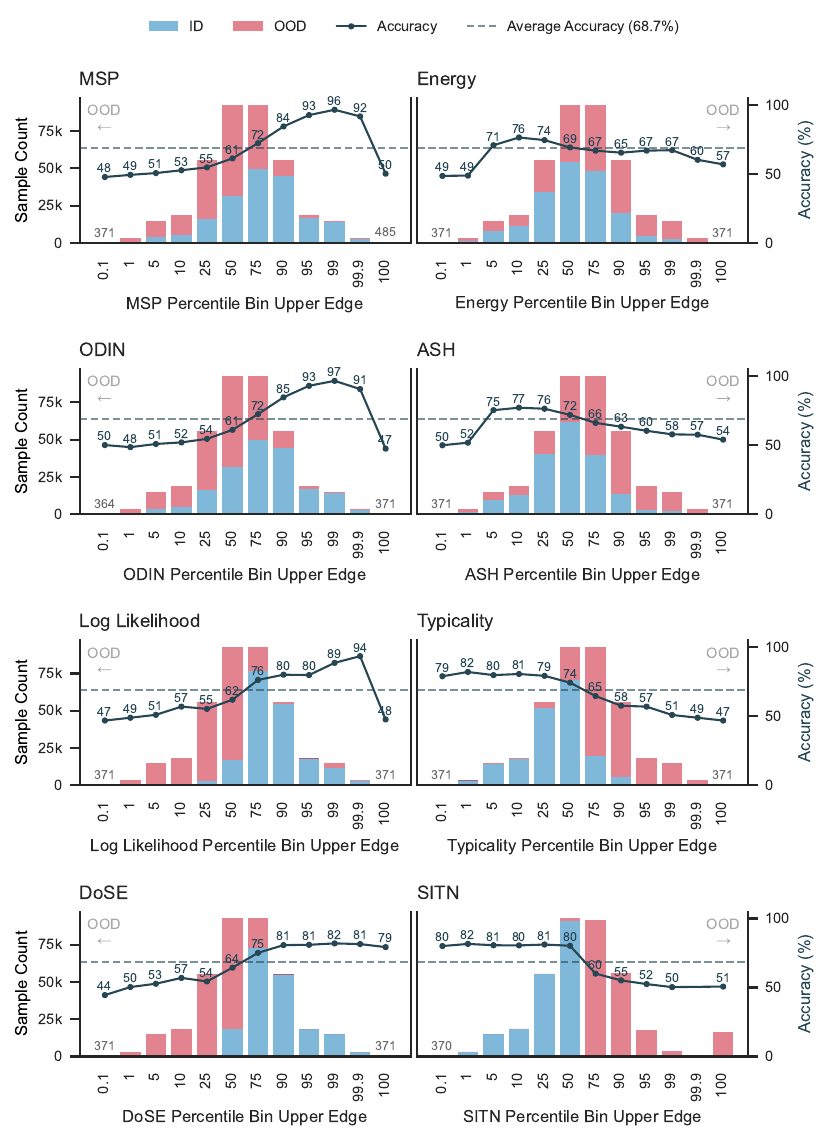}
            \caption{
                \textbf{Normality Prediction on TUAB --- Re-referencing Perturbation (bipolar)} \\
                Note: The irregular bin sizes at the upper percentiles of the SITN plot are due to metric saturation, where a large number of tied maximum scores are grouped into the final bin.
            }
        \end{figure}

        \begin{figure}[H]
            \centering
            \includegraphics[width=\textwidth]{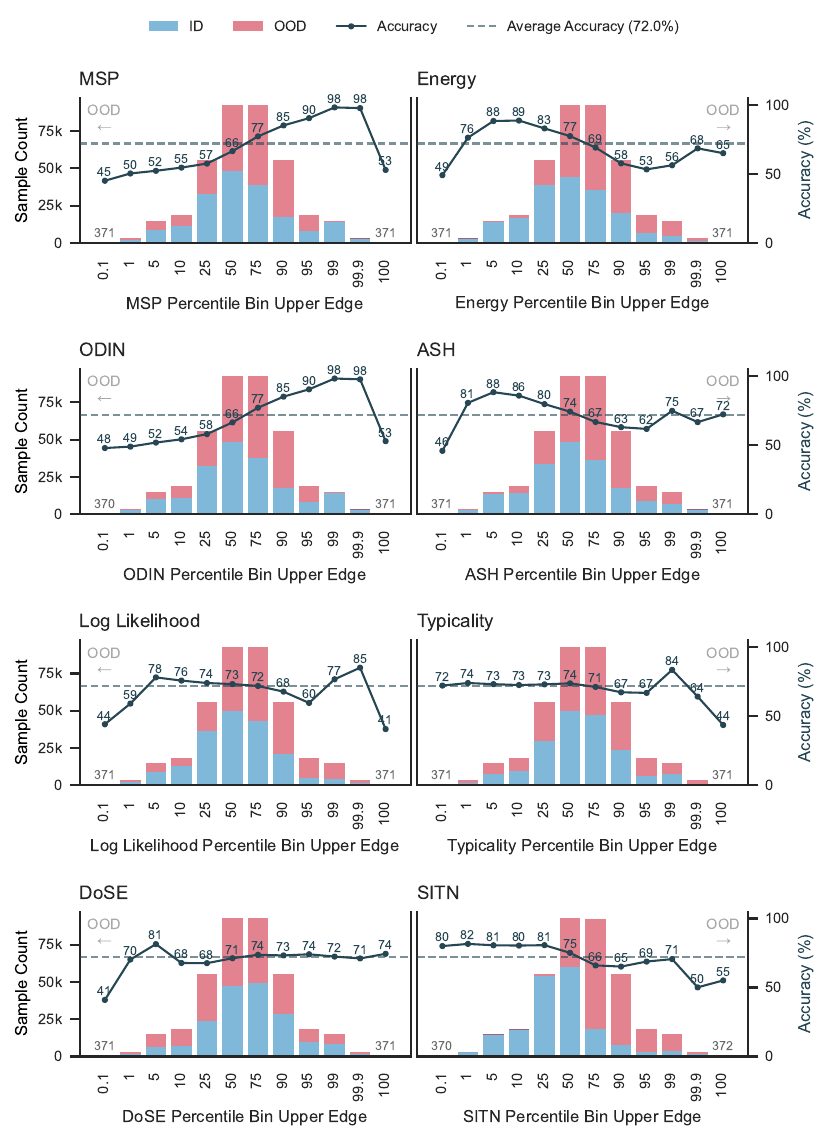}
            \caption{
                \textbf{Normality Prediction on TUAB --- High-pass Filter Perturbation (8 Hz)}
            }
        \end{figure}

        \begin{figure}[H]
            \centering
            \includegraphics[width=\textwidth]{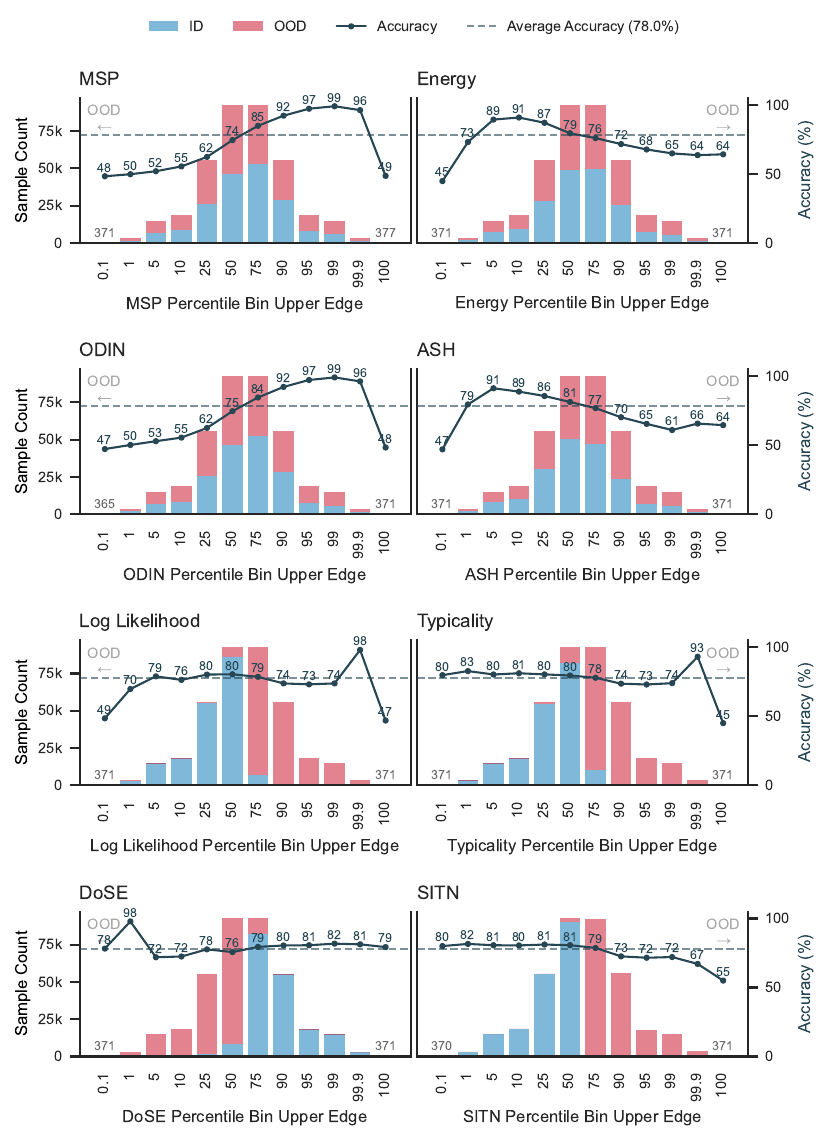}
            \caption{
                \textbf{Normality Prediction on TUAB --- Low-pass Filter Perturbation (15 Hz)}
            }
        \end{figure}

        \begin{figure}[H]
            \centering
            \includegraphics[width=\textwidth]{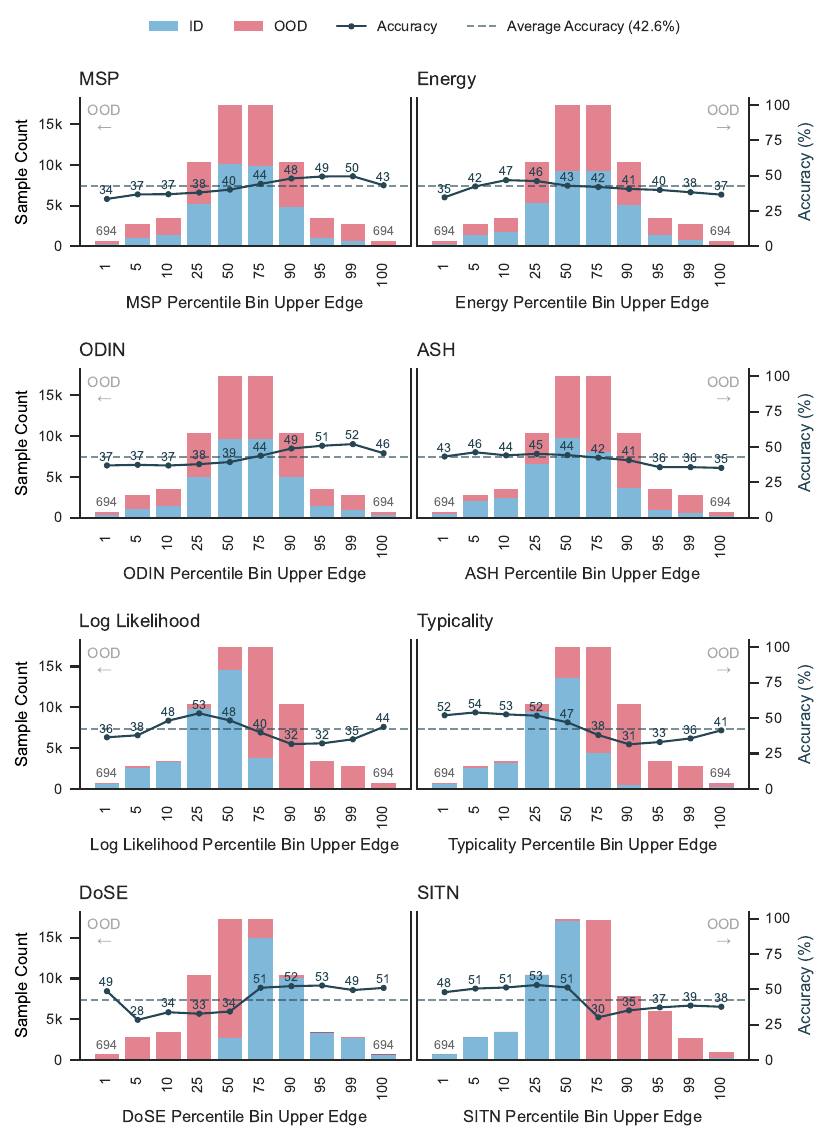}
            \caption{
                \textbf{Dementia Diagnosis on CAUEEG --- Resampling Perturbation (200 Hz)}
            }
        \end{figure}

        \begin{figure}[H]
            \centering
            \includegraphics[width=\textwidth]{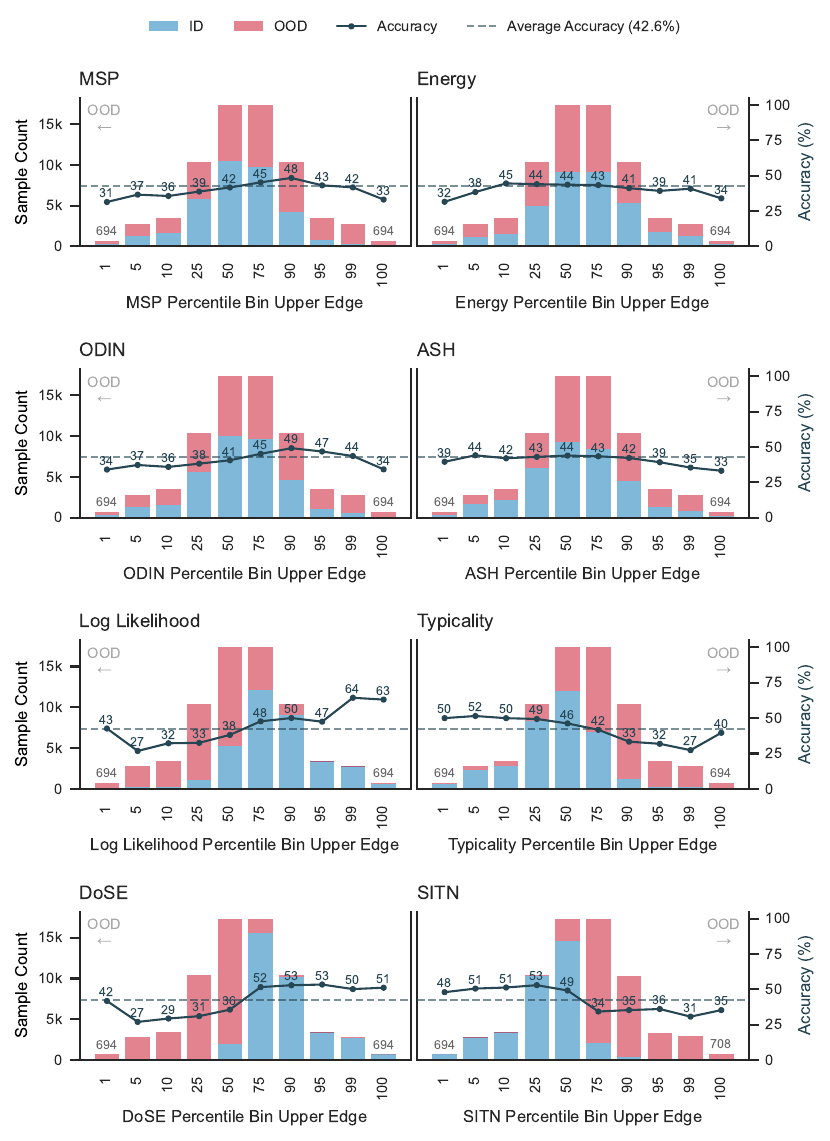}
            \caption{
                \textbf{Dementia Diagnosis on CAUEEG --- Channel Shuffle Perturbation (100\%)}
            }
        \end{figure}

        \begin{figure}[H]
            \centering
            \includegraphics[width=\textwidth]{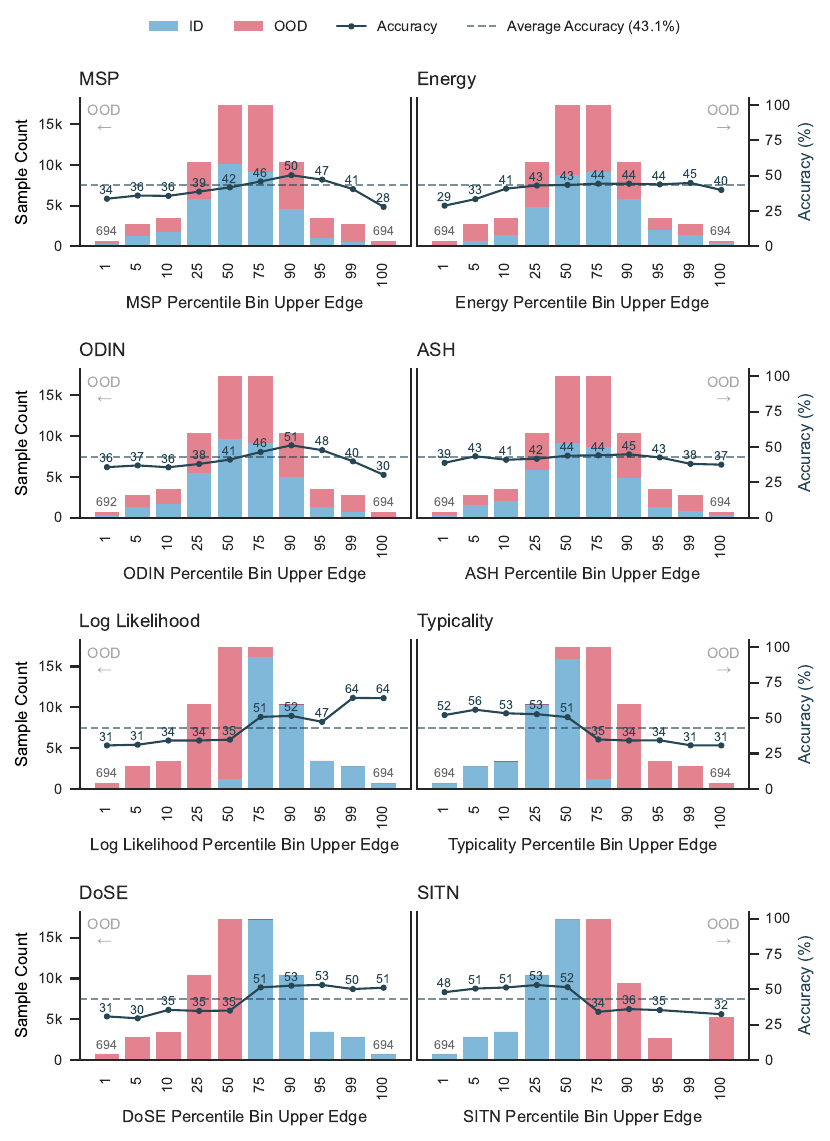}
            \caption{
                \textbf{Dementia Diagnosis on CAUEEG --- Re-referencing Perturbation (bipolar)} \\
                Note: The irregular bin sizes at the upper percentiles of the SITN plot are due to metric saturation, where a large number of tied maximum scores are grouped into the final bin.
            }
        \end{figure}

        \begin{figure}[H]
            \centering
            \includegraphics[width=\textwidth]{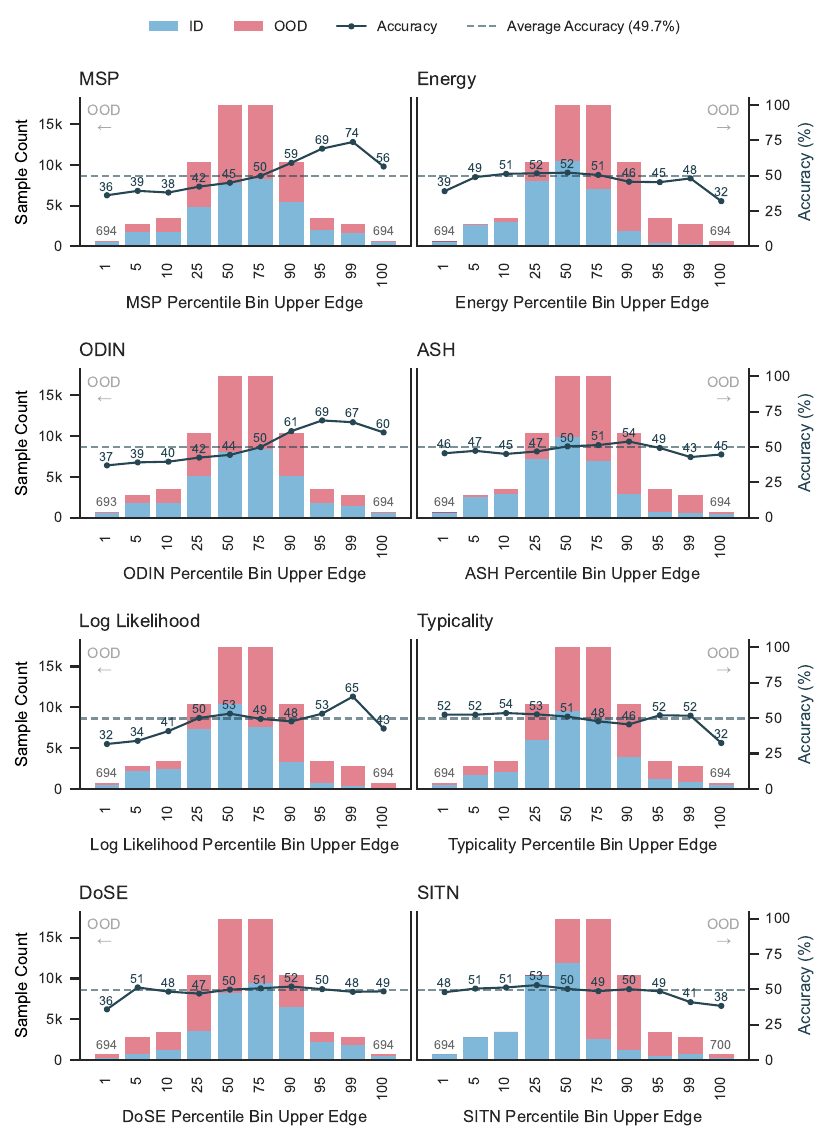}
            \caption{
                \textbf{Dementia Diagnosis on CAUEEG --- High-pass Filter Perturbation (8 Hz)}
            }
        \end{figure}

        \begin{figure}[H]
            \centering
            \includegraphics[width=\textwidth]{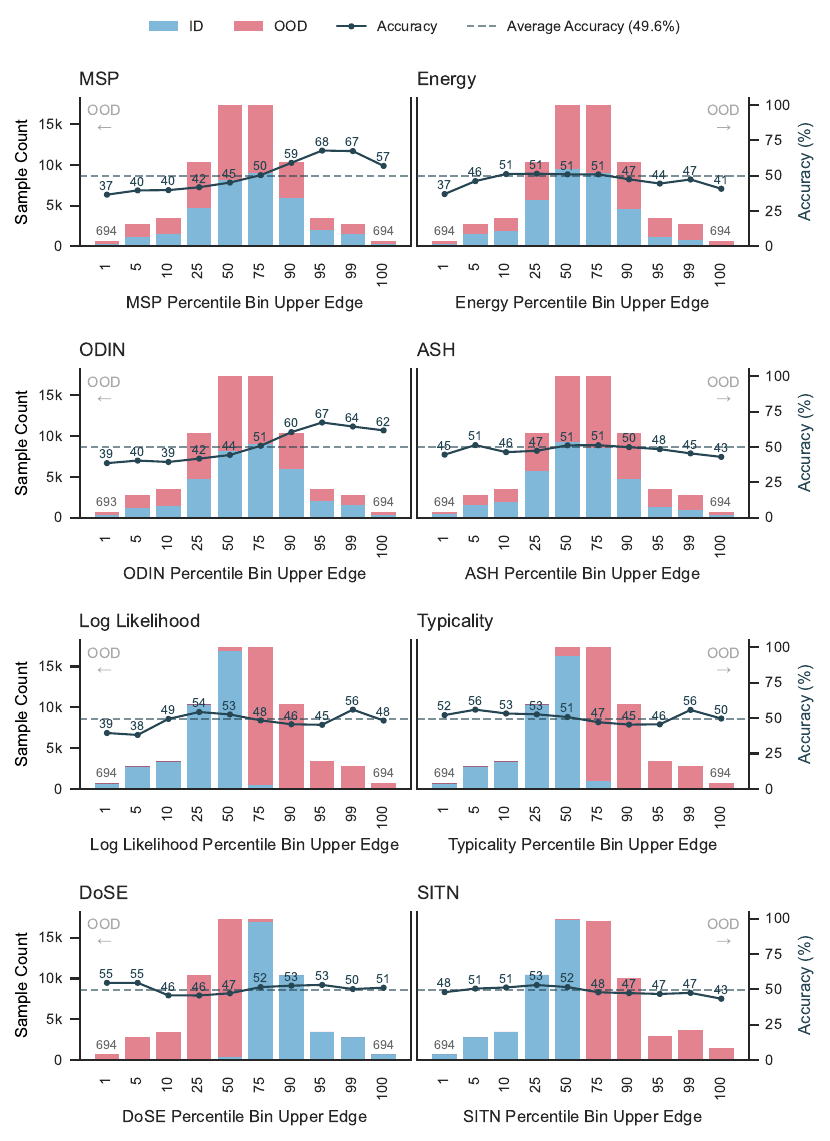}
            \caption{
                \textbf{Dementia Diagnosis on CAUEEG --- Low-pass Filter Perturbation (15 Hz)}
            }
            \label{fig:appendix_downstream_lowpass_caueeg}
        \end{figure}

\end{appendices}

\end{document}